\documentclass[final,authoryear,3p,times]{elsarticle}

\usepackage{afterpage}
\usepackage{algorithm}
\usepackage{algpseudocode}
\usepackage{amsmath}
\usepackage{amssymb}
\usepackage{amsthm}
\usepackage{bm}
\usepackage{booktabs}
\usepackage{calc}
\usepackage{caption}
\usepackage{subcaption}
\usepackage[table]{xcolor} 
\usepackage{amsmath,amssymb}
\usepackage{booktabs}
\usepackage{tabularx}
\usepackage{adjustbox}

\usepackage{enumerate}
\usepackage{graphicx}
\usepackage{ifthen}
\usepackage[parfill]{parskip} 
\usepackage{multicol}
\usepackage{siunitx}
\usepackage{ulem}

\usepackage[colorlinks]{hyperref}
\usepackage[nameinlink,capitalise]{cleveref}
\crefname{appendix}{}{}

\usepackage{tikz}
\usetikzlibrary{arrows}
\usetikzlibrary{automata}
\usetikzlibrary{calc}
\usetikzlibrary{chains}
\usetikzlibrary{positioning}
\usetikzlibrary{shapes}

\usepackage{varwidth}

\usepackage{tcolorbox}
\newtcolorbox[auto counter]{problem}[2][]{colframe=blue!30, colback=blue!5, coltitle=black, title=Problem~\thetcbcounter~ #2,#1}

\usepackage{spverbatim} 
\usepackage{listings} 

\definecolor{Demiray}{rgb}{0.1725, 0.6275, 0.1725}
\definecolor{Gent}{rgb}{1.0, 0.4902, 0.0431}
\definecolor{Holzapfel}{rgb}{0.1216, 0.4667, 0.7059}
\definecolor{MooneyRivlin}{rgb}{0.8392, 0.1529, 0.1569}
\definecolor{Ogden}{rgb}{0.5804, 0.4039, 0.7412}
\definecolor{BlatzKo}{rgb}{0.5490, 0.3373, 0.2941}
\definecolor{NeoHooke}{rgb}{0.7373, 0.7412, 0.1333}

\usepackage{cleveref}
\usepackage{makecell}

\edef\svtheparindent{\the\parindent}
\usepackage{parskip}
\usepackage{ulem}
\parindent=\svtheparindent\relax

\definecolor{ForestGreen}{RGB}{34,139,34}
\definecolor{InternationalOrange}{rgb}{1.0, 0.31, 0.0}
\definecolor{WineRed}{RGB}{139,0,0}

\newcommand{\boldface}[1]{\boldsymbol{#1}}  

\newcommand{\bfb}{\boldface{b}}

\newcommand{\bfq}{\boldface{q}}

\newcommand{\bfx}{\boldface{x}}

\newcommand{\bfz}{\boldface{z}}
\newcommand{\bfA}{\boldface{A}}

\newcommand{\bfC}{\boldface{C}}

\newcommand{\bfF}{\boldface{F}}

\newcommand{\bfI}{\boldface{I}}

\newcommand{\bfP}{\boldface{P}}
\newcommand{\bfQ}{\boldface{Q}}

\newcommand{\bfU}{\boldface{U}}

\newcommand{\bfW}{\boldface{W}}

\newcommand{\bftheta}{\boldsymbol{\theta}}

\newcommand{\bfzero}{\boldsymbol{0}}
\newcommand{\Rset}{\mathbb{R}}

\newcommand{\be}{\begin{equation}}
\newcommand{\ee}{\end{equation}}
\newcommand{\bea}{\begin{equation}\begin{aligned}}
\newcommand{\eea}{\end{aligned}\end{equation}}
\newcommand{\beq}{\begin{eqnarray}}
\newcommand{\eeq}{\end{eqnarray}}
\newcommand{\bem}{\begin{multline}}
\newcommand{\eem}{\end{multline}}
\newcommand{\ba}{\begin{align}}
\newcommand{\ea}{\end{align}}
\newcommand{\bcase}{\left\{ \begin{array}{ll}}
\newcommand{\ecase}{\end{array} \right.}
\newcommand{\shortto}{\mathrel{\scalebox{0.5}[1]{$\rightarrow$}}}

\begin{document}

\begin{frontmatter}

\title{
Benchmarking data-driven material models on the classic Treloar dataset
}


\cortext[cor1]{Correspondence: hagen.holthusen@fau.de}
\author[fau]{Hagen Holthusen\corref{cor1}}
\author[fau]{Moritz Flaschel}
\author[fau]{Denisa Martonová}
\author[fau,stan]{Ellen Kuhl}
\address[fau]{Institute of Applied Mechanics, Egerlandstraße 5, Friedrich-Alexander-Universität Erlangen-Nürnberg, 91058 Erlangen, Germany}
\address[stan]{Department of Mechanical Engineering, Stanford University, 440 Escondido Mall, California 94305, United States.}

\begin{abstract}
Machine learning is rapidly reshaping constitutive modeling, offers new ways to learn material behavior directly from experimental data, and challenges long-established modeling paradigms. But with a growing number of machine-learning-based approaches available, how do they compare in practice?
In this paper, we use the classic experimental data of \textbf{Treloar} to benchmark popular frameworks for hyperelasticity:
\textbf{(Generalized-Invariant) Constitutive Artificial Neural Networks}, \textbf{Physics-Augmented Neural Networks}, \textbf{(Adaptive) Material Fingerprinting}, and \textbf{Efficient Unsupervised Constitutive Law Identification \& Discovery}.
We compare their fitting performance, computational cost, hyperparameter sensitivity, and ease of implementation. Furthermore, we discuss the trade-offs between predictive accuracy and model complexity. The latter is assessed by quantifying both the number of material parameters in the discovered models and the computational time required to evaluate the constitutive model and its derivatives. The results show that all methods can reproduce the benchmark data remarkably well. Rather than identifying a single winner, we highlight the strengths and limitations of each approach and provide practical guidance for their use.
The source code for all six methods, including the training and comparison scripts, as well as all results and data used in this study, is publicly available via \href{https://doi.org/10.5281/zenodo.21915635}{Zenodo}.

\end{abstract}


\end{frontmatter}



\section{Introduction}

\begin{figure*}[t]
  \centering
  \includegraphics[width=0.4\textwidth]{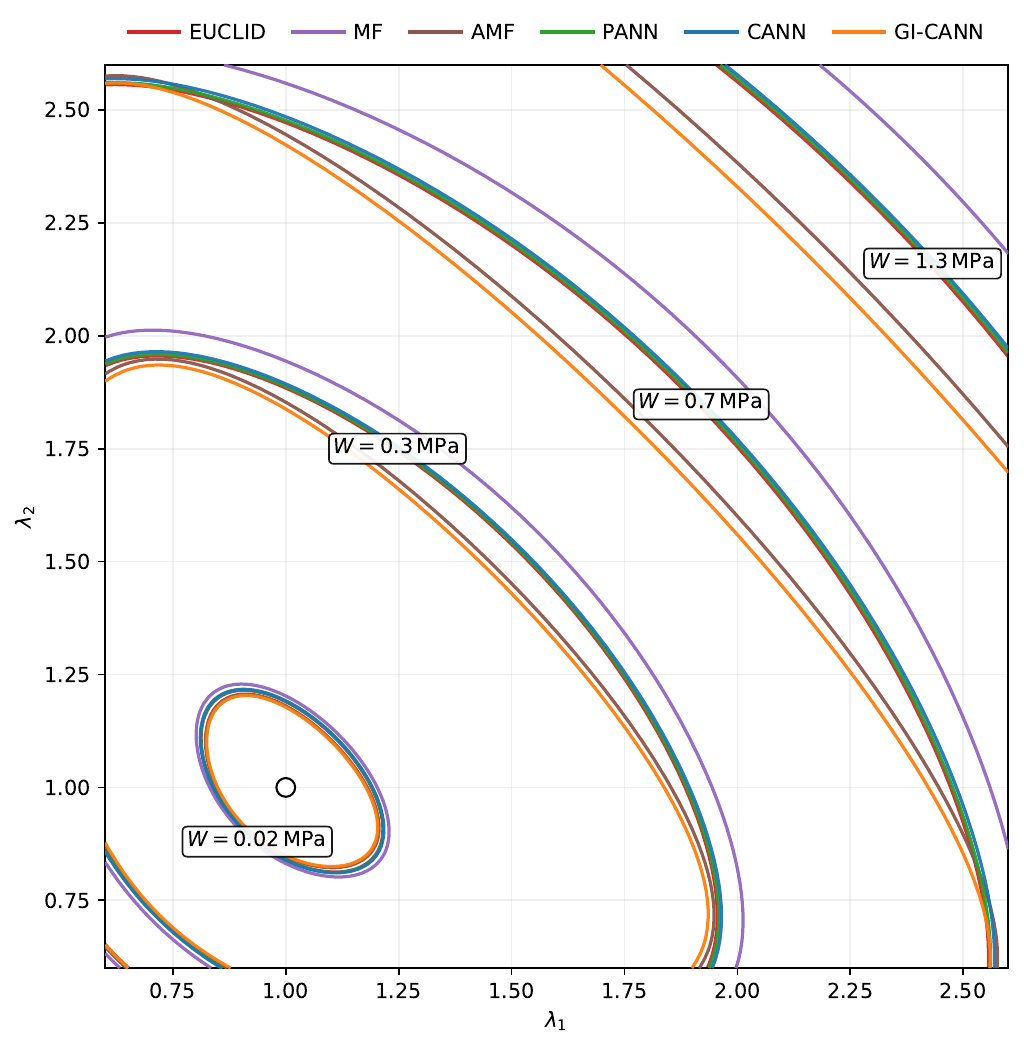}
  \caption{
  Strain energy density in the principal stretch space predicted by the six selected best-performing models (see \cref{sec:Results} for details).
  }
  \label{fig:strain_energy_isolines_overlay}
\end{figure*}

\begin{figure*}[t]
  \centering
  \includegraphics[width=\textwidth]{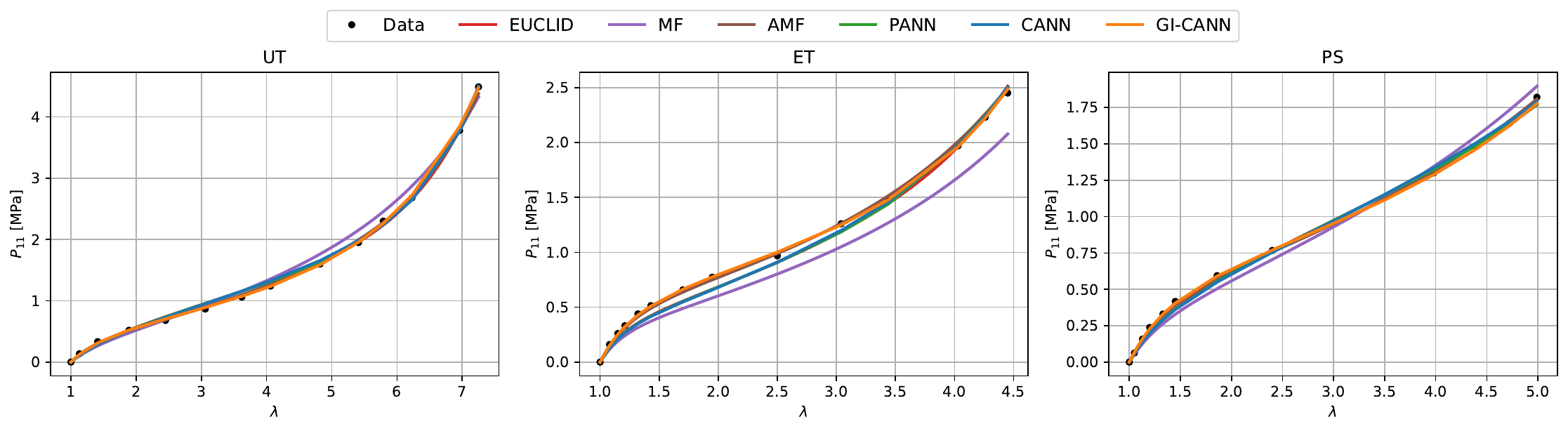}
  \caption{
  Comparison of Treloar's experimental data and predictions from the best-performing discovered models.
  }
  \label{fig:best_models_comparison}
\end{figure*}

Treloar's experiments on natural rubber \citep{treloar_stress-strain_1944} constitute one of the most widely used benchmarks for the development and evaluation of constitutive models. A systematic classical comparison assessed fourteen phenomenological and micromechanically motivated hyperelastic models against these data and examined whether parameters identified from one deformation mode could predict the others \citep{steinmann_hyperelastic_2012}. More than ten years later, the emergence of machine learning and automated model discovery methods has since changed the modeling landscape: rather than selecting exclusively from a limited set of prescribed constitutive forms, these approaches can learn or assemble the strain energy density function from data, with different balances of physical structure, flexibility, sparsity, and interpretability. This shift introduces new practical questions concerning generalization, optimization robustness, hyperparameter and initialization sensitivity, and computational cost. A renewed benchmark on the same established dataset is therefore timely. Here, we compare six representative data-driven approaches under common conditions. An overview of the considered methods is provided in \cref{tab:results_overview}. \cref{fig:strain_energy_isolines_overlay} shows their discovered constitutive functions, and \cref{fig:best_models_comparison} shows their model response in comparison with Treloar's data.

\begin{table*}[h]
  \centering
  \caption{
  Benchmark performance of data-driven methods
  (see \cref{sec:Results} and \cref{tab:best_models} for details).
  }
  \label{tab:results_overview}
  \small
  \setlength{\tabcolsep}{4.5pt}
  \begin{adjustbox}{max width=\textwidth}
  \begin{tabular}{@{}llrrrr@{}}
    \toprule
    Method
      & Acronym
      & \makecell{Training\\$R^2$}
      & \makecell{Validation\\$R^2$}
      & \makecell{Core time\\(ms)}
      & \makecell{\# Parameters\\$\|\boldsymbol{\theta}\|_0$}
    \\
    \midrule
    Efficient Unsupervised Constitutive Law Identification \& Discovery
      & EUCLID
      & 0.9973
      & 0.9978
      & 0.372
      & 4
      \\
    Material Fingerprinting
      & MF
      & 0.9781
      & 0.9899
      & \textbf{0.302}
      & \textbf{2}
      \\
    Adaptive Material Fingerprinting
      & AMF
      & 0.9991
      & \textbf{0.9996}
      & 6.000
      & 49
      \\
    Physics-Augmented Neural Networks
      & PANN
      & 0.9973
      & 0.9977
      & 581.311
      & 371
      \\
    Constitutive Artificial Neural Networks
      & CANN
      & 0.9972
      & 0.9975
      & 320.947
      & 12
      \\
    Generalized Invariant-based CANN
      & GI-CANN
      & \textbf{0.9996}
      & 0.9993
      & 664.988
      & 8
      \\
    \bottomrule
  \end{tabular}%
  \end{adjustbox}
\end{table*}

To quantitatively assess the performance of the considered methods on Treloar's dataset, we introduce several metrics that evaluate the fitting accuracy of the resulting models. In addition to fitting accuracy, we investigate the computational cost of the considered data-driven methods and evaluate the complexity of the resulting constitutive models using several complementary metrics. Constitutive model complexity is particularly relevant for computational simulations, where the repeated evaluation of the model and its derivatives at every integration point can have a significant impact on the overall computational cost. \cref{tab:results_overview} summarizes some representative metrics used to quantify fitting accuracy, model complexity, and the computational cost of the inverse problem. Additional performance measures will be introduced and discussed throughout this work.

We note that the literature on data-driven constitutive modeling is extensive and continues to grow rapidly. Consequently, our comparison is restricted to a representative set of widely used approaches and does not claim to cover all state-of-the-art methods or identify the universally best-performing techniques. Specifically, we consider Efficient Unsupervised Constitutive Law Identification and Discovery (EUCLID) \citep{flaschel_automated_2023-1}, Material Fingerprinting (MF) \citep{flaschel_material_2026,martonova_2026a}, Adaptive Material Fingerprinting (AMF) \citep{flaschel_adaptive_2026}, Physics-Augmented Neural Networks (PANNs) \citep{klein_polyconvex_2022,linden_neural_2023,dammass_2025a}, Constitutive Artificial Neural Networks (CANNs) \citep{linka_constitutive_2021,linka_new_2023}, and Generalized Invariant-based CANNs (GI-CANNs) \citep{martonova_2026}. This list of methods is by no means exhaustive. Numerous other approaches have been proposed in the literature \citep{fuhg_review_2024,tac_benchmarking_2024}, including, to name only a few, Gaussian processes \citep{frankel_tensor_2020,fuhg_local_2022}, Variational System Identification (VSI) \citep{wang_inference_2021}, Neural Ordinary Differential Equations (NODE) \citep{tac_data-driven_2022}, symbolic regression \citep{abdusalamov_automatic_2023}, spline-based approaches \citep{wiesheier_versatile_2024}, non-smooth parameterizations of constitutive functions \citep{bleyer_learning_2025}, and agentic artificial intelligence \citep{tacke_constitutive_2025}.

In the following \cref{sec:Benchmark}, we briefly review the classical Treloar dataset on natural rubber. In the subsequent \cref{sec:Methods}, we introduce the considered data-driven methods for constitutive modeling. Next, we apply these methods to the benchmark dataset and analyze and discuss the results in \cref{sec:Results}.

\section{Benchmark problem}
\label{sec:Benchmark}

To benchmark different data-driven material modeling approaches with respect to predictive accuracy and computational efficiency, we apply several widely used methods to the same dataset under identical conditions. Before introducing the individual methods, we first describe the benchmark dataset and its preprocessing, the training and test data split, the training objective employed by the data-driven models, and the evaluation metrics used to quantify predictive performance.

\subsection{Dataset and preprocessing}

In this work, we use the classical Treloar rubber dataset at $20^\circ$C as a benchmark for data-driven constitutive model discovery \citep{treloar_stress-strain_1944}. The dataset comprises stress--stretch measurements obtained from three canonical homogeneous deformation modes: uniaxial tension (UT), equibiaxial tension (ET), and pure shear (PS).

The experimental data are given as stretch values $\lambda$ and corresponding nominal first Piola-Kirchhoff stress values $P_{11}$. Since the constitutive models operate on the full three-dimensional deformation gradient, each scalar stretch value is first converted into a three-dimensional deformation gradient $\bfF$. Assuming incompressibility, the deformation gradients are constructed as
\begin{equation}
\bfF^{\mathrm{UT}}
=
\mathrm{diag}
\left(
\lambda,
\lambda^{-1/2},
\lambda^{-1/2}
\right), \quad
\bfF^{\mathrm{ET}}
=
\mathrm{diag}
\left(
\lambda,
\lambda,
\lambda^{-2}
\right), \quad
\bfF^{\mathrm{PS}}
=
\mathrm{diag}
\left(
\lambda,
1,
\lambda^{-1}
\right),
\end{equation}
for uniaxial tension, equibiaxial tension, and pure shear, respectively. For each experiment, a set of stretch--stress data tuples is available, see \cref{fig:best_models_comparison},
\begin{equation}
    \mathcal{D}^{\mathrm{UT}} = \left\{ (\lambda_i^{\mathrm{UT}}, P_{11,i}^{\mathrm{UT}}) \, | \, i = 1,\dots, N_{\mathrm{UT}} \right\}, \quad
    \mathcal{D}^{\mathrm{ET}} = \left\{ (\lambda_i^{\mathrm{ET}}, P_{11,i}^{\mathrm{ET}}) \, | \, i = 1,\dots, N_{\mathrm{ET}} \right\}, \quad
    \mathcal{D}^{\mathrm{PS}} = \left\{ (\lambda_i^{\mathrm{PS}}, P_{11,i}^{\mathrm{PS}}) \, | \, i = 1,\dots, N_{\mathrm{PS}} \right\}.    
\end{equation}
For training, the uniaxial and equibiaxial datasets are concatenated into one calibration dataset, and the pure shear data are used for testing
\begin{equation}
\mathcal{D}_{\mathrm{train}}
=
\mathcal{D}^{\mathrm{UT}}
\cup
\mathcal{D}^{\mathrm{ET}}, \quad
\mathcal{D}_{\mathrm{test}}
=
\mathcal{D}^{\mathrm{PS}}.
\end{equation}
The pure shear data are not used during training and are retained as an independent validation case. This allows us to assess whether the discovered constitutive model generalizes to a deformation mode that was not part of the calibration data.

To improve numerical stability during optimization, the nominal stress targets are normalized by the maximum absolute stress value in the training set
\begin{equation}
\widehat{P}_{11,i}
=
\frac{P_{11,i}}{\max\limits_{j=1,\dots,N_{\mathrm{train}}} |P_{11,j}|}.
\end{equation}
The data-driven models are trained using the normalized stress values. During evaluation, the predicted stresses are transformed back to physical units using the corresponding normalization factor. No smoothing, filtering, or data augmentation is applied, such that the comparison remains directly based on the original experimental measurements.

\subsection{Training objective}

All data-driven methods considered in this work seek to identify a parametric representation of the constitutive model. In the context of hyperelasticity, this corresponds to learning a strain energy density function $W_{\boldsymbol{\theta}}$ parameterized by the model parameters $\boldsymbol{\theta}$. The parameters are learned in a supervised manner by minimizing the discrepancy between the model predictions and the experimentally measured nominal stresses. For each deformation gradient $\bfF_i$ in the training dataset, the model predicts the first Piola--Kirchhoff stress from the learned strain energy density
\begin{equation}
\bfP_{\boldsymbol{\theta}}(\bfF_i)
=
\frac{\partial W_{\boldsymbol{\theta}}(\bfF_i)}{\partial \bfF_i}.
\end{equation}
For incompressible materials, the stress is corrected by a pressure-like Lagrange multiplier so that the transverse stress condition is satisfied. In the present implementation, this pressure correction is chosen from the condition $P_{33} = 0$
which removes the hydrostatic contribution from the predicted first Piola--Kirchhoff stress. The loss is therefore evaluated on the physically admissible incompressible stress response rather than on the unconstrained energy gradient.

Since the experimental data provide the nominal stress component in the loading direction, the loss is formulated on the normalized $P_{11}$ component. The training objective is the mean squared error
\begin{equation}
\mathcal{L}_{\mathrm{data}}(\boldsymbol{\theta})
=
\frac{1}{N_{\mathrm{train}}}
\sum_{i=1}^{N_{\mathrm{train}}}
\left(
\widehat{P}_{11,\boldsymbol{\theta}}(\bfF_i)
-
\widehat{P}_{11,i} 
\right)^2 ,
\end{equation}
where $\widehat{P}_{11,i}$ denotes the normalized experimental stress and $\widehat{P}_{11,\boldsymbol{\theta}}$ denotes the corresponding model prediction normalized by the same factor as the experimental stress.

The total optimization problem is given by
\begin{equation}
\boldsymbol{\theta}^\ast
=
\arg\min_{\boldsymbol{\theta}}
\mathcal{L}_{\mathrm{data}}(\boldsymbol{\theta}),
\end{equation}
possibly supplemented by architecture-dependent constraints or regularization terms that enforce constitutive structure, positivity, convexity, or sparsity. Finally, we note that not all data-driven methods considered in this work are formulated as continuous optimization problems of the form above. In particular, database-driven Material Fingerprinting methods instead rely on pattern-recognition algorithms operating on a pre-established material database.

\subsection{Evaluation metrics}

In this work, we compare different data-driven methods based on several evaluation criteria. These include the computational time required to identify or discover them from the given data, the goodness of fit of the identified models, and the interpretability or simplicity of the resulting models. We introduce different metrics for quantifying the goodness of fit. For a given dataset $\mathcal{D}$ with a number of $|\mathcal{D}|$ data points, we consider the mean squared error (MSE)
\begin{equation}
\text{MSE}_{\mathcal{D}}
=
\frac{1}{|\mathcal{D}|}
\sum_{i=1}^{|\mathcal{D}|}
\left(
{P}_{11,\boldsymbol{\theta}}(\lambda_i)
-
{P}_{11,i}
\right)^2 ,
\end{equation}
the root mean squared error (RMSE)
\begin{equation}
\text{RMSE}_{\mathcal{D}}
=
\sqrt{
\frac{1}{|\mathcal{D}|}
\sum_{i=1}^{|\mathcal{D}|}
\left(
{P}_{11,\boldsymbol{\theta}}(\lambda_i)
-
{P}_{11,i}
\right)^2} ,
\end{equation}
the range-normalized root mean squared error (NRMSE)
\begin{equation}
\text{NRMSE}_{\mathcal{D}}
=
\frac{\text{RMSE}_{\mathcal{D}}}{\max\limits_{j=1,\dots,|\mathcal{D}|} |P_{11,j}| - \min\limits_{j=1,\dots,|\mathcal{D}|} |P_{11,j}|}
 ,
\end{equation}
the mean absolute error
\begin{equation}
\text{MAE}_{\mathcal{D}}
=
\frac{1}{|\mathcal{D}|}
\sum_{i=1}^{|\mathcal{D}|}
\left|
{P}_{11,\boldsymbol{\theta}}(\lambda_i)
-
{P}_{11,i}
\right| ,
\end{equation}
and the coefficient of determination
\begin{equation}
R^2_{\mathcal{D}} \;=\; 1 - \frac{\sum_{i=1}^{|\mathcal{D}|} \left( {P}_{11,\boldsymbol{\theta}}(\lambda_i) - {P}_{11,i} \right)^2}{\sum_{i=1}^{|\mathcal{D}|} \left( \text{mean}_{\mathcal{D}} ({P}_{11,i}) - {P}_{11,i} \right)^2},
\end{equation}
where $\text{mean}_{\mathcal{D}} ({P}_{11,i})$ is the arithmetic mean of all experimental stress values in the dataset.

In constitutive modeling, it is often the case that multiple models provide an equally accurate description of the available data, while differing substantially in their complexity, interpretability, and computational cost. For practical applications, particularly in large-scale computational simulations, simpler constitutive models are generally preferred, as they require less computational effort to evaluate the strain energy density function and its derivatives. Consequently, in addition to predictive accuracy, model complexity constitutes an important criterion for assessing and comparing constitutive models. We consider different metrics to quantify the model complexity of the identified models. 

The first measure of model complexity considered in this work is the effective number of material parameters (or trainable weights in the terminology of machine learning). Specifically, we count only those parameters that remain nonzero after the model identification procedure, as these determine the complexity of the resulting constitutive model. We denote this quantity by $\|\boldsymbol{\theta}\|_0$, in analogy with the $L_0$-pseudo-norm, which counts the number of nonzero entries in a vector.

While the number of material parameters \(\|\boldsymbol{\theta}\|_0\) provides a useful proxy for model complexity, it does not directly reflect the computational cost of evaluating a strain energy density function and its derivatives. Constitutive models with similar numbers of parameters may require substantially different numbers and types of operations, resulting in different computational costs when deployed in finite element simulations. To quantify this aspect, we implemented each identified strain energy density function in a common machine learning framework and used automatic differentiation to evaluate the energy $W$, its gradient $\bfP$, and its Hessian $\mathbb{H}$. Each benchmark evaluated a batch of \(N_{\mathrm{eval}}=100\) deformation gradients in double precision on the CPU. After two warm-up evaluations, the computation was repeated 100 times. The total runtime of each repetition was divided by \(N_{\mathrm{eval}}\), yielding the runtime per deformation gradient. Compilation, parameter loading, sample generation, validation, and host result transfer were excluded. The resulting energy, gradient, and Hessian evaluation times are reported as the mean \(\pm\) standard deviation over the 100 repetitions.

\section{Methods}
\label{sec:Methods}

In the following, we briefly introduce the data-driven material modeling approaches considered in this work. Detailed descriptions of the individual methods can be found in the corresponding original publications. An overview of the considered approaches is provided in \cref{tab:hyperparameters_overview}, which summarizes the optimization algorithms and solvers employed for model identification from experimental data, together with the associated hyperparameters and modeling choices. These hyperparameters and modeling choices are discussed in detail for each method in the subsequent subsections.

\begin{table}[ht]
\centering
\caption{Hyperparameters of machine learning methods for constitutive modeling.}
\scriptsize\renewcommand{\arraystretch}{1.4}
\begin{tabularx}{\textwidth}{
>{\raggedright\arraybackslash}X
>{\raggedright\arraybackslash}p{4.cm}
>{\raggedright\arraybackslash}p{4.cm}
>{\raggedright\arraybackslash}p{4.cm}
}
\toprule
\textbf{Method} & \textbf{Solvers / Algorithms} & \textbf{Hyperparameters} & \textbf{Modeling Decisions} \\
\midrule

EUCLID
& $\bullet$ Coordinate Descent (CD) \newline
$\bullet$ Least Angle Regression (LARS)
& $\bullet$ Number of terms in the library \newline
$\bullet$ Regularization parameter $\alpha$ \newline
$\bullet$ Thresholding parameter $\theta_{{\shortto}{0}}$ 
& $\bullet$ Choice of the library features \newline
$\bullet$ Model input variables
\\

\hline
Material Fingerprinting
& $\bullet$ Pattern-recognition 
& $\bullet$ Number of fingerprints in the database \newline
$\bullet$ Parameter sampling during database generation
& $\bullet$ Choice of the models in the database \newline
$\bullet$ Model input variables
\\

\hline
Adaptive Material Fingerprinting
& $\bullet$ Adaptive pattern-recognition 
& $\bullet$ Number of fingerprints in the database \newline
$\bullet$ Parameter sampling during database generation \newline
$\bullet$ Number of terms in the model $N_a$ \newline
$\bullet$ Step size $s\in [0,1]$
& $\bullet$ Choice of the modeling features in the database \newline
$\bullet$ Model input variables
\\

\hline
PANN
& Various (usually gradient-based) \newline
$\bullet$ SGD \newline
$\bullet$ Adam \newline
$\bullet$ L-BFGS
& $\bullet$ Learning rate $\eta$ \newline
$\bullet$ Number of epochs \newline
$\bullet$ Number of layers \newline
$\bullet$ Number of neurons per layer \newline
$\bullet$ Optional ML hyperparameters (e.g., regularization, early stopping, dropout, scheduling)
& $\bullet$ Choice of the activation functions \newline
$\bullet$ Model input variables
\\

\hline
CANN and GI-CANN
& Various (usually gradient-based) \newline
$\bullet$ SGD \newline
$\bullet$ Adam \newline
$\bullet$ L-BFGS
& $\bullet$ Learning rate $\eta$ \newline
$\bullet$ Number of epochs \newline
$\bullet$ Number of layers \newline
$\bullet$ Number of neurons per layer \newline
$\bullet$ Optional ML hyperparameters (e.g., regularization, early stopping, dropout, scheduling)
& $\bullet$ Choice of the activation functions \newline
$\bullet$ Model input variables
\\

\bottomrule
\end{tabularx}
\label{tab:hyperparameters_overview}
\end{table}

For the benchmark problem considered in this work, all data-driven methods share the common objective of constructing a mathematical representation of the hyperelastic strain energy density function. In the present study, we restrict our attention to isotropic incompressible hyperelasticity, for which the strain energy density function can be expressed as
\begin{equation}
    \tilde{W}(\bfF) = W(\bfF) - p[J-1],
\end{equation}
where \(J=\det(\bfF)\) and \(p\) is a scalar Lagrange multiplier enforcing the incompressibility constraint \(J=1\). To ensure material isotropy, the constitutive contribution \(W(\bfF)\) is commonly expressed in terms of invariants of the right Cauchy--Green deformation tensor \(\bfC=\bfF^{T}\bfF\). The most common choices are the first and second principal invariants,
\(I_1=\operatorname{tr}(\bfC)\) and
\(I_2=\frac{1}{2}\left[(\operatorname{tr}\bfC)^2-\operatorname{tr}(\bfC^2)\right]\),
or the principal stretches \(\lambda_1,\lambda_2,\lambda_3\), whose squares are the eigenvalues of \(\bfC\), i.e.,
\(\{\lambda_1^2,\lambda_2^2,\lambda_3^2\}=\operatorname{eig}(\bfC)\).
These quantities satisfy
\(I_1=\lambda_1^2+\lambda_2^2+\lambda_3^2\),
\(I_2=\lambda_1^2\lambda_2^2+\lambda_2^2\lambda_3^2+\lambda_3^2\lambda_1^2\), and
\(J=\lambda_1\lambda_2\lambda_3=1\).
Alternatively, generalized invariant sets may be employed, as discussed below for the GI-CANN framework.

By differentiating \(\tilde{W}\) with respect to the deformation gradient and determining \(p\) from the constraint \(P_{33}=0\) for the dataset introduced in \cref{sec:Benchmark}, we obtain the \(P_{11}\) component of the first Piola--Kirchhoff stress as
\begin{equation}
    P_{11}
    =
    \frac{\partial \tilde{W}}{\partial F_{11}}
    =
    \frac{\partial W}{\partial F_{11}}
    -
    \frac{\partial W}{\partial F_{33}}
    \left[\operatorname{cof}(\bfF)_{33}\right]^{-1}
    \operatorname{cof}(\bfF)_{11}.
\end{equation}

The following sections describe how EUCLID, Material Fingerprinting, Adaptive Material Fingerprinting, PANNs, CANNs, and GI-CANNs identify, discover, or learn the constitutive contribution \(W\).

\subsection{Efficient Unsupervised Constitutive Law Identification and Discovery (EUCLID)}

EUCLID is a method for the automated discovery of interpretable material models with a small number of modeling terms from data \citep{flaschel_unsupervised_2021}. The underlying idea is to construct a library or catalog of modeling terms and use sparse regression to identify the combination of terms that best describes the given data. In its original form \citep{flaschel_unsupervised_2021}, EUCLID is unsupervised, which means that it does not necessarily rely on stress-stretch data pairs but can instead be informed by full-field displacement measurements of complexly shaped specimens. However, the concepts of EUCLID can equally be applied in a supervised fashion when stress-stretch data pairs are available \citep{flaschel_automated_2023-1}, as is the case for Treloar's data. In the following, we demonstrate how EUCLID can be applied to Treloar's data to efficiently discover material models.

At the heart of EUCLID stands a set of candidate modeling terms that could potentially describe the material behavior. In the case of isotropic hyperelasticity, this can be, for example, the terms of the generalized Mooney-Rivlin model, which are motivated by a Taylor expansion of the invariant-based strain energy density about the undeformed configuration. We collect these modeling terms in the feature vector
\begin{equation}
    \bfQ(I_1,I_2) = \left[ [I_1-3] , \, [I_2-3], \, [I_1-3]^2, \, [I_1-3][I_2-3], \, [I_2-3]^2, \, \dots  \right]^T.
\end{equation}
In this work, we consider coupled polynomial features up to an order of five, resulting in $N_{\text{feat}}=20$ modeling features. We note, however, that the library can be readily extended with more modeling features, such as those of the Ogden material model \citep{flaschel_automated_2023-1}.

A linear combination of the candidate modeling terms yields the strain energy density
\begin{equation}
\label{eq:EUCLID_W}
    W(I_1,I_2) = \bfQ(I_1,I_2) \cdot \bftheta,
\end{equation}
where $\bftheta \in \Rset_{\geq 0}^{N_{\text{feat}}}$ are material parameters, which we assume here to be non-negative. 

The objective of EUCLID is to identify the parameters $\bftheta$ such that model predictions are similar to the experimental measurements while, at the same time, removing candidate modeling terms that are irrelevant for describing the data. To this end, we consider the regularized optimization problem 
\begin{equation}
\label{eq:EUCLID_optimization}
\bftheta^\ast
=
\arg\min_{\bftheta\in\Rset_{\geq 0}}
\mathcal{L}_{\mathrm{data}}(\bftheta) + \alpha \| \bftheta \|_1,
\end{equation}
in which the $L_1$-norm penalizes solutions with many non-zero parameters and promotes sparsity in the solution \citep{frank_statistical_1993,tibshirani_regression_1996,hastie_elements_2009,brunton_discovering_2016}. The regularization parameter $\alpha \geq 0$ determines how aggressively the parameters should be pushed to zero. Since our model is linear in the parameters $\bftheta$ (see \cref{eq:EUCLID_W}), the loss term in the optimization problem above is quadratic in $\bftheta$. The $L_1$-norm regularization term is non-smooth but convex in $\bftheta$. The total objective function is convex and non-smooth and can be efficiently minimized using methods such as Coordinate Descent (CD) \citep{fu_penalized_1998} or Least Angle Regression (LARS) \citep{efron_least_2004}. Such non-smooth optimization approaches have also been applied to constitutive model discovery and material modeling \citep{flaschel_non-smooth_2026}.

Solving the problem above yields a parameter vector $\bftheta$ with many zero entries and, hence, a material model that is expressed by a small number of modeling terms. We note that some parameters may not be exactly zero but may instead attain values very close to zero. In such cases, a threshold parameter $\theta_{{\shortto}{0}}$  can be introduced in a post-processing step, whereby parameters with absolute values below the threshold are set to zero \citep{flaschel_unsupervised_2021}. For the results presented in this paper, however, such a thresholding step was not necessary to obtain a sparse representation of the model and was therefore omitted. Furthermore, once EUCLID has identified a suitable sparse set of constitutive terms, it can be beneficial to perform a final regression with $\alpha=0$ to optimize only the remaining nonzero parameters. This post-processing step removes the bias introduced by the regularization and can further improve model accuracy, but it has not been employed in the present work.

\begin{tcolorbox}[title=Hyperparameters]
As shown in \cref{tab:hyperparameters_overview}, EUCLID depends on several hyperparameters, including the number of terms in the candidate library, the regularization parameter $\alpha$, and the threshold parameter $\theta_{{\shortto}{0}}$. In addition, its performance is influenced by several modeling choices made during the construction of the library. In particular, the variables on which the library features depend (here, the principal invariants), as well as the functional forms included in the library must be specified a priori. We finally note that previous works on EUCLID have also investigated the nonconvex $L_p$-regularization term with $0<p<1$ \citep{flaschel_unsupervised_2021,mcculloch_sparse_2024}. However, since increasing the regularization parameter $\alpha$ allows the $L_1$-norm to achieve any desired level of sparsity, we restrict our attention here to the convex $L_1$-regularization.
\end{tcolorbox}




\subsection{Material Fingerprinting (MF)}

Material Fingerprinting is a database-driven method for constitutive model discovery that replaces the solution of a continuous optimization problem with a pattern-recognition task \citep{flaschel_material_2026,martonova_2026a}. The key assumption is that every material exhibits a characteristic mechanical response when subjected to a standardized experimental protocol. This response defines the material's fingerprint. Once a sufficiently rich database of fingerprints exists, material characterization reduces to identifying the closest fingerprint in the database.

Material Fingerprinting consists of an offline and an online stage. During the offline stage, we generate a database of fingerprints by numerically simulating standardized experiments for different constitutive models (Blatz-Ko, Demiray, Gent, Holzapfel, Mooney-Rivlin, Neo-Hooke, Ogden) and parameter combinations. Each constitutive model is represented by its specific strain energy density function $W$ and sampled over a prescribed range of material parameters \citep{martonova_2026a}. In the supervised setting considered here, the experiments correspond to homogeneous deformation modes, such as uniaxial tension and simple shear, and each fingerprint collects the stresses at prescribed deformation states. More generally, the method also applies to heterogeneous experiments that provide full-field displacement and reaction force measurements \citep{flaschel_unsupervised_2026}. Each database entry stores a fingerprint together with its constitutive model and material parameters
\begin{equation}
    \left(
    \mathbf f^{(i)},
    \boldsymbol{\theta}^{(i)}
    \right),
    \qquad i=1,\ldots,N_{\mathrm{db}},
\end{equation}
where $N_{\mathrm{db}}$ denotes the number of fingerprints in the database.

To remove the dependence on the overall stiffness scale, we normalize every fingerprint
\begin{equation}
    \bar{\mathbf f}^{(i)}
    =
    \frac{\mathbf f^{(i)}}{\|\mathbf f^{(i)}\|},
\end{equation}
which exploits the homogeneity of many hyperelastic constitutive models with respect to their material parameters. As a result, materials that differ only by a multiplicative scaling of their parameters share the same normalized fingerprint, which substantially reduces the required database size \citep{flaschel_material_2026}.

During the online stage, we measure and normalize the fingerprint $\mathbf f^\ast$ of an unknown material. We then compare the measured fingerprint with every database entry through the cosine similarity
\begin{equation}
\label{eq:MF_cosine}
    \text{cs}^{(i)}
    =
    \frac{
        \bar{\mathbf f}^{(i)}
        \cdot
        \bar{\mathbf f}^{\ast}
    }{
        \left|
        \bar{\mathbf f}^{(i)}
        \right|
        \left|
        \bar{\mathbf f}^{\ast}
        \right|
    }
    =
    \bar{\mathbf f}^{(i)}
    \cdot
    \bar{\mathbf f}^{\ast},
\end{equation}
where the second equality follows because the fingerprints have unit norm. Equivalently, Material Fingerprinting identifies the fingerprint that forms the smallest angle with the experimental fingerprint. We therefore identify the constitutive model through
\begin{equation}
    i^\ast
    =
    \arg\max_i
    \;
    \text{cs}^{(i)}.
\end{equation}

Unlike conventional parameter identification methods, Material Fingerprinting does not solve a continuous optimization problem during the online stage. Instead, it reduces constitutive model discovery to a matrix--vector multiplication followed by a maximum search.  Moreover, because the algorithm searches the complete database, it identifies the discrete global optimum within the searchable model space instead of converging to a local optimum of a continuous optimization problem \citep{flaschel_material_2026}. Consequently, the method shifts the computational effort entirely to the offline stage, while the online characterization of new materials requires only milliseconds. In this work, we focus on supervised Material Fingerprinting for experiments with homogeneous deformation fields. We note that the computational speed-up is even more pronounced in the unsupervised case for experiments with heterogeneous deformation fields \citep{flaschel_unsupervised_2026}.


\begin{tcolorbox}[title=Hyperparameters]
As shown in \cref{tab:hyperparameters_overview}, the performance of Material Fingerprinting depends on several user-defined modeling choices and hyperparameters. For instance, the variables on which the constitutive models in the database are based (here, the principal invariants and principal stretches), as well as the admissible functional forms, must be specified a priori. In addition, the database generation requires selecting the sampling density of the material parameters, which determines the total number of fingerprints contained in the database. These hyperparameter choices affect only the database generation and therefore constitute a one-time effort. Once an optimized and sufficiently expressive database has been constructed, it can be reused repeatedly for efficient material model discovery.
\end{tcolorbox}

\subsection{Adaptive Material Fingerprinting (AMF)}

Adaptive Material Fingerprinting extends the original Material Fingerprinting framework by constructing the constitutive model incrementally instead of selecting a single predefined model from the database \citep{flaschel_adaptive_2026}. Whereas the original method stores fingerprints of complete constitutive models, the adaptive method stores fingerprints of individual modeling features. It reconstructs the constitutive response step by step by successively adding the feature that best explains the discrepancy between the experimental data and the current model prediction.

During the offline stage, we generate a database of feature fingerprints instead of fingerprints for complete constitutive models. We express the strain energy density as a linear combination of modeling features,
\begin{equation}
    W(\bfF)
    =
    \sum_{a=1}^{N_a}
    \theta_a
    \,
    Q_a(\bfF),
\end{equation}
where $N_a$ denotes the maximum number of feature terms included in the discovered constitutive model. We compute the stress response associated with every feature for different parameter choices. The feature library may contain isotropic and anisotropic terms based on principal stretches, principal area changes, or fiber stretches. This representation captures constitutive models such as the multi-term Ogden model or the Holzapfel-Gasser-Ogden model \citep{flaschel_adaptive_2026}. Because we prescribe the admissible parameter ranges during database generation, we can also enforce physical constraints, such as polyconvexity, by restricting the database to admissible feature combinations.

During the online stage, we initialize the predicted fingerprint with
\begin{equation}
    \mathbf f_0=\mathbf0.
\end{equation}
At iteration $k$, we compute the residual fingerprint
\begin{equation}
    \mathbf r_k
    =
    \mathbf f^\ast
    -
    \mathbf f_k,
\end{equation}
where $\mathbf f^\ast$ denotes the experimental fingerprint and $\mathbf f_k$ denotes the fingerprint predicted by the current constitutive model. We then compare the normalized residual with all feature fingerprints in the database through the cosine similarity
\begin{equation}
    i_k^\ast
    =
    \arg\max_i
    \;
    \frac{
        \bar{\mathbf f}^{(i)}
        \cdot
        \bar{\mathbf r}_k
    }{
        \left|
        \bar{\mathbf f}^{(i)}
        \right|
        \left|
        \bar{\mathbf r}_k
        \right|
    }.
\end{equation}

After we identify the best matching feature, we add it to the constitutive model. Its contribution is scaled by a step size parameter $s\in(0,1]$, which controls how much of the current residual is incorporated at each iteration. Thus, smaller values of $s$ lead to more gradual updates, whereas $s=1$ corresponds to incorporating the full residual contribution. We repeat the procedure for a prescribed maximum number of iterations $N_a$. 

Compared to the original Material Fingerprinting framework, the adaptive method increases the flexibility of the constitutive model because it combines multiple modeling features instead of selecting a single predefined model. At the same time, it retains the computational efficiency of database-based pattern recognition and, dependent on the choice of $N_a$, produces sparse, physically interpretable constitutive models without solving a continuous optimization problem \citep{flaschel_adaptive_2026}.


\begin{tcolorbox}[title=Hyperparameters]
As shown in \cref{tab:hyperparameters_overview}, Adaptive Material Fingerprinting depends on a larger number of hyperparameters than the original Material Fingerprinting method. In particular, the number of terms $N_a$ included in the model and the step size $s$ must be specified a priori. Due to the computational efficiency of Adaptive Material Fingerprinting, suitable values for these hyperparameters can be determined using a simple grid search \citep{flaschel_adaptive_2026}.
\end{tcolorbox}

\subsection{Physics-Augmented Neural Networks (PANNs)}

Physics-Augmented Neural Networks \citep{klein_polyconvex_2022,linden_neural_2023} incorporate constitutive requirements directly into the neural network architecture. For the isotropic incompressible setting considered here, we represent the strain energy density by an input-convex neural network (ICNN) \citep{amos_input_2017} that takes the first and second principal invariants as inputs,
\begin{equation}
    W(I_1,I_2)
    =
    \operatorname{ICNN}_{\bftheta}(I_1,I_2)
    -
    \operatorname{ICNN}_{\bftheta}(3,3),
\end{equation}
where \(\bftheta\) denotes the trainable network parameters. Since \(I_1=I_2=3\) in the undeformed reference configuration, the second term normalizes the strain energy density such that \(W(3,3)=0\). This term is constant with respect to the deformation and therefore does not affect the predicted stresses or tangent operators. Moreover, subtracting a constant preserves the convexity and monotonicity properties of the ICNN. {The additional stress normalization according to \citet{linden_neural_2023} is not required at this point, as the Lagrange multiplier $p$ ensures a stress-free undeformed reference  configuration.}

The choice of an ICNN is motivated by the construction of a polyconvex strain energy density. In three dimensions, the principal invariants can be written as
\begin{equation}
    I_1
    =
    \bfF:\bfF,
    \qquad
    I_2
    =
    \operatorname{cof}(\bfF):
    \operatorname{cof}(\bfF).
\label{eq:I1_I2_in_F}
\end{equation}
Thus, \(I_1\) is a convex function of \(\bfF\), while \(I_2\) is a convex function of \(\operatorname{cof}(\bfF)\). If $W(I_1,I_2)$ is convex and non-decreasing in each argument, the composition is jointly convex in \(\bfF\) and \(\operatorname{cof}(\bfF)\), regarded as independent arguments. Consequently, a strain energy density of the form
\begin{equation}
    W(\bfF)
    =
    \operatorname{ICNN}_{\bftheta}
    \left(
        \bfF:\bfF,\,
        \operatorname{cof}(\bfF):
        \operatorname{cof}(\bfF)
    \right)
    -
    \operatorname{ICNN}_{\bftheta}(3,3)
\end{equation}
is polyconvex if the ICNN is convex and componentwise non-decreasing in its inputs. For incompressible deformations, \(I_1\geq3\) and \(I_2\geq3\). Componentwise monotonicity, together with the normalization at \((I_1,I_2)=(3,3)\), therefore additionally ensures \(W\geq0\) for all admissible deformation states.

To describe how these properties are embedded into the network, we collect the invariants in the input vector
\begin{equation}
    \bfx
    =
    \begin{bmatrix}
        I_1, & I_2
    \end{bmatrix}^{T}.
\end{equation}
The first hidden representation is defined as
\begin{equation}
    \bfz_1
    =
    \varphi_1
    \left(
        \bfU_1\bfx+\bfb_1
    \right),
\end{equation}
and the subsequent representations are constructed recursively as
\begin{equation}
    \bfz_{\ell}
    =
    \varphi_{\ell}
    \left(
        \bfW_{\ell}\bfz_{\ell-1}
        +
        \bfU_{\ell}\bfx
        +
        \bfb_{\ell}
    \right),
    \qquad
    \ell=2,\ldots,L.
\end{equation}
The final representation is scalar and defines the network output,
\begin{equation}
    \operatorname{ICNN}_{\bftheta}(\bfx)
    =
    \bfz_L.
\end{equation}
The ICNN can therefore be understood as a composition of the layer mappings
\begin{equation}
    g_{\ell}(\bfz;\bfx)
    =
    \varphi_{\ell}
    \left(
        \bfW_{\ell}\bfz
        +
        \bfU_{\ell}\bfx
        +
        \bfb_{\ell}
    \right).
\end{equation}
Importantly, each layer depends not only on the preceding hidden representation but also directly on the original input through the term \(\bfU_{\ell}\bfx\). These direct input paths constitute the skip connections of the ICNN architecture.

Convexity of the ICNN with respect to \(\bfx\) is ensured by choosing activation functions \(\varphi_{\ell}\) that are convex and non-decreasing and by imposing the componentwise constraints
\begin{equation}
    \bfW_{\ell}\geq\bfzero,
    \qquad
    \ell=2,\ldots,L.
\end{equation}
Indeed, the non-negative hidden-to-hidden weights preserve convexity when the representations from the preceding layer are combined, while the skip-connection term \(\bfU_{\ell}\bfx\) is affine in \(\bfx\). Input convexity alone would therefore not require a sign constraint on \(\bfU_{\ell}\).
For the polyconvex construction above, however, the ICNN must additionally be non-decreasing in both \(I_1\) and \(I_2\). This is ensured by also constraining the input and skip-connection weights componentwise according to
\begin{equation}
    \bfU_{\ell}\geq\bfzero,
    \qquad
    \ell=1,\ldots,L.
\end{equation}
Together with the non-negative hidden-to-hidden weights and the non-decreasing activation functions, these constraints make every layer, and hence the complete network, componentwise non-decreasing in \(\bfx\). The biases \(\bfb_{\ell}\) affect neither convexity nor monotonicity and remain unconstrained.

Although the ICNN is convex with respect to its invariant inputs and the resulting strain energy density is polyconvex, the loss function is generally non-convex with respect to the network parameters. The identified model may therefore depend on their initialization.

\begin{tcolorbox}[title=Hyperparameters]
As shown in \cref{tab:hyperparameters_overview}, the design and training of PANNs depend on several user-defined modeling choices and hyperparameters. Prior to training, users must specify the network architecture, including the number of hidden layers, the number of neurons per layer, the activation functions, and the input variables on which the network depends. The training process is subsequently governed by hyperparameters such as the learning rate and the number of training epochs, as well as optional techniques including regularization, early stopping, dropout, and learning rate scheduling.
\end{tcolorbox}

\subsection{Constitutive Artificial Neural Networks (CANNs)}

Constitutive Artificial Neural Networks \citep{linka_constitutive_2021,linka_new_2023} represent the strain energy density by a structured neural network whose architecture resembles a generalized constitutive model. Rather than learning an unrestricted mapping from deformation measures to energy, CANNs combine physically motivated invariant-based features through trainable activation functions and non-negative weights \citep{holthusen_theory_2024}.

For the isotropic incompressible setting considered here, the network takes the shifted first and second principal invariants
\begin{equation}
    \bar{I}_1=I_1-3,
    \qquad
    \bar{I}_2=I_2-3
\end{equation}
as inputs. Since \(I_1\geq3\) and \(I_2\geq3\) for incompressible deformations, both shifted invariants are non-negative over the admissible deformation domain. We construct the basic feature vector
\begin{equation}
    \bfq(\bar{I}_1,\bar{I}_2)
    =
    \begin{bmatrix}
        \bar{I}_1, &
        \bar{I}_2, &
        \bar{I}_1^2, &
        \bar{I}_2^2
    \end{bmatrix}^{T}.
\end{equation}
Each feature enters the strain energy density both directly and through an exponential activation.\footnote{For numerical stability, the arguments of the exponential activations are clipped to a finite interval in the implementation. This safeguard was introduced because the exponential branches acting on the quadratic features, of the form \(\exp(w_{1,k}^{\mathrm{exp}}\bar{I}_i^2)\), can otherwise overflow during training. The clipping is not part of the constitutive ansatz; the strict convexity argument applies over the range in which the clipping remains inactive.} The resulting CANN can be written as
\begin{equation}
    W(I_1,I_2)
    =
    \sum_{k=1}^{4}
    w_{2,k}^{\mathrm{id}}\,q_k
    +
    \sum_{k=1}^{4}
    w_{2,k}^{\mathrm{exp}}
    \left[
        \exp\left(
            w_{1,k}^{\mathrm{exp}}\,q_k
        \right)-1
    \right],
\end{equation}
where \(w_{2,k}^{\mathrm{id}}\) are the output weights of the identity branches, \(w_{1,k}^{\mathrm{exp}}\) control the exponential activations, and \(w_{2,k}^{\mathrm{exp}}\) are the corresponding exponential output weights. Hence, the CANN comprises a total of 12 trainable weights. This construction ensures that
\begin{equation}
    W(3,3)=0
\end{equation}
is satisfied by construction, without requiring an additional energy normalization.

All trainable weights are constrained componentwise according to
\begin{equation}
    w_{2,k}^{\mathrm{id}}\geq0,
    \qquad
    w_{1,k}^{\mathrm{exp}}\geq0,
    \qquad
    w_{2,k}^{\mathrm{exp}}\geq0,
    \qquad
    k=1,\ldots,4.
\end{equation}
These constraints ensure that each constitutive branch contributes non-negatively to the strain energy density. Since the shifted invariants are non-negative, the linear and quadratic features \(q_k\) are convex and non-decreasing functions of their respective invariant. For \(w_{1,k}^{\mathrm{exp}}\geq0\), the exponential mapping
\begin{equation}
    q_k
    \mapsto
    \exp\left(
        w_{1,k}^{\mathrm{exp}}\,q_k
    \right)-1
\end{equation}
is likewise convex and non-decreasing. A non-negative weighted sum of these terms therefore remains convex and componentwise non-decreasing in \(I_1\) and \(I_2\).

As discussed for the PANN framework, these properties provide a sufficient construction for polyconvexity. In three dimensions, the invariants can be expressed in terms of \(\bfF\) and \(\operatorname{cof}(\bfF)\) according to \cref{eq:I1_I2_in_F} \citep{holthusen2026complement}. Because the CANN is convex and non-decreasing in both invariants, its composition with these convex functions admits a jointly convex representation in \(\bfF\) and \(\operatorname{cof}(\bfF)\) \citep{holthusen2026generalized}. The resulting strain energy density is therefore polyconvex over the admissible incompressible deformation domain. Together with the invariant-based inputs, the architecture also ensures objectivity and material isotropy.

\begin{tcolorbox}[title=Hyperparameters]
As shown in \cref{tab:hyperparameters_overview}, the hyperparameters of CANNs are largely similar to those of the PANNs discussed earlier. The primary difference is that CANNs incorporate handcrafted and expert-guided functional forms directly into the network architecture.
\end{tcolorbox}

\subsection{Generalized-Invariant-based Constitutive Artificial Neural Networks (GI-CANNs)}

Generalized-Invariant-based Constitutive Artificial Neural Networks \citep{martonova_2026} extend the CANN framework by treating not only the functional form of the strain energy density, but also its invariant representation as trainable. Instead of restricting the model input to the classical invariants \(I_1\) and \(I_2\), GI-CANNs introduce the continuous family of generalized invariants \citep{anssari-benam_generalised_2024}
\begin{equation}
    \mathcal{J}_{\alpha}
    =
    \sum_{i=1}^{3}
    \lambda_i^{\alpha},
\end{equation}
where \(\lambda_i\) are the principal stretches and \(\alpha\in\mathbb{R}\) is an exponent that can be identified from the data. For incompressible materials, the classical invariants are recovered as the special cases
\begin{equation}
    \mathcal{J}_{2}=I_1,
    \qquad
    \mathcal{J}_{-2}=I_2.
\end{equation}
The trainable exponent therefore allows the network to continuously explore invariant representations beyond the fixed classical choices.

In the general GI-CANN framework, the strain energy density depends on a set of generalized invariants,
\begin{equation}
    W
    =
    W\left(
        \left\{
            \mathcal{J}_{\alpha_k}
        \right\}_{\alpha_k\in\mathcal{S}}
    \right),
\end{equation}
where \(\mathcal{S}\subset\mathbb{R}\) denotes the set of invariant exponents. Fixing \(\mathcal{S}=\{2,-2\}\) recovers the classical invariant representation underlying the standard-invariant-based CANN \citep{linka_new_2023}. In contrast, prescribing a discrete grid of exponents and combining the resulting generalized invariants through identity branches recovers the principal-stretch-based CANN \citep{pierre_principalstretchbased_2023}. GI-CANN generalizes both approaches by treating the exponents as continuous trainable parameters and identifying the invariant representation and the strain energy density function simultaneously.

For the benchmark implementation considered here, the exponent set contains one non-negative exponent and one non-positive exponent
\begin{equation}
    \mathcal{S}
    =
    \left\{
        \alpha_{+},-\alpha_{-}
    \right\},
    \qquad
    \alpha_{+}\geq0,
    \quad
    \alpha_{-}\geq0.
\end{equation}
The corresponding shifted generalized invariants are
\begin{equation}
    \bar{\mathcal{J}}_{+}
    =
    \sum_{i=1}^{3}
    \lambda_i^{\alpha_{+}}-3,
    \qquad
    \bar{\mathcal{J}}_{-}
    =
    \sum_{i=1}^{3}
    \lambda_i^{-\alpha_{-}}-3.
\end{equation}
In the numerical implementation, the principal stretches are computed as the singular values of the deformation gradient. Subtracting three ensures that both shifted generalized invariants vanish in the undeformed reference configuration.

The general GI-CANN architecture may apply different powers and nonlinear transformations to the shifted generalized invariants. In the present implementation, we restrict the power set to the identity and use one linear and one exponential branch for each invariant. The strain energy density is therefore expressed as
\begin{equation}
    W
    =
    \sum_{k\in\{+,-\}}
    \left\{
        w_{2,k}^{\mathrm{id}}\,
        \bar{\mathcal{J}}_{k}
        +
        w_{2,k}^{\mathrm{exp}}
        \left[
            \exp\left(
                w_{1,k}^{\mathrm{exp}}
                \bar{\mathcal{J}}_{k}
            \right)-1
        \right]
    \right\}.
\end{equation}
Thus, the model simultaneously identifies two generalized invariant exponents, two weights controlling the exponential activations, and four output weights, resulting in a total of eight trainable parameters.

All trainable parameters are constrained componentwise according to
\begin{equation}
    \alpha_{+}\geq0,
    \qquad
    \alpha_{-}\geq0,
    \qquad
    w_{1,k}^{\mathrm{exp}}\geq0,
    \qquad
    w_{2,k}^{\mathrm{id}}\geq0,
    \qquad
    w_{2,k}^{\mathrm{exp}}\geq0.
\end{equation}
For an incompressible deformation, the product of the principal stretches satisfies \(\lambda_1\lambda_2\lambda_3=1\). Consequently, \(\mathcal{J}_{\alpha}\geq3\) for any real exponent \(\alpha\), and both shifted generalized invariants are non-negative. Together with the non-negative network weights, this ensures that every constitutive branch contributes non-negatively and that
\begin{equation}
    W\geq0,
    \qquad
    W(1,1,1)=0.
\end{equation}
Because the principal stretches enter through symmetric sums, the resulting strain energy density is objective and isotropic. Together with the incompressibility pressure correction, the undeformed configuration is stress-free.

Noteworthy, the GI-CANN architecture admits a sufficient condition for polyconvexity in the cosidered incompressible case. For \(\alpha_{+}\geq1\), the generalized invariant
\begin{equation}
    \mathcal{J}_{\alpha_{+}}
    =
    \sum_{i=1}^{3}
    \sigma_i(\bfF)^{\alpha_{+}}
\end{equation}
is a convex spectral function of \(\bfF\), where \(\sigma_i(\bfF)=\lambda_i\) are its singular values. Under the incompressibility constraint, the inverse principal stretches coincide with the singular values of the cofactor. Consequently,
\begin{equation}
    \mathcal{J}_{-\alpha_{-}}
    =
    \sum_{i=1}^{3}
    \lambda_i^{-\alpha_{-}}
    =
    \sum_{i=1}^{3}
    \sigma_i\left(
        \operatorname{cof}(\bfF)
    \right)^{\alpha_{-}}
\end{equation}
is a convex spectral function of \(\operatorname{cof}(\bfF)\) for \(\alpha_{-}\geq1\). Since the subsequent CANN mapping is convex and componentwise non-decreasing in both generalized invariants, the resulting strain energy density then admits a jointly convex representation in \(\bfF\) and \(\operatorname{cof}(\bfF)\), and is therefore polyconvex.

In the present implementation, the exponent magnitudes are constrained only by \(\alpha_{+},\alpha_{-}\geq0\). Polyconvexity is therefore not guaranteed a priori if an active constitutive branch attains an exponent between zero and one. Imposing the stronger constraints
\begin{equation}
    \alpha_{+}\geq1,
    \qquad
    \alpha_{-}\geq1
\end{equation}
would provide a sufficient architectural guarantee of polyconvexity. These conditions only need to be satisfied by active constitutive branches; the exponent associated with a branch whose output weights vanish does not affect the resulting strain energy density.

\begin{tcolorbox}[title=Hyperparameters]
As shown in \cref{tab:hyperparameters_overview}, the hyperparameters of GI-CANNs are equivalent to those of the CANNs discussed earlier.
\end{tcolorbox}

\section{Results}
\label{sec:Results}

All computations use the same software implementation and parameter
settings across the six methods. Detailed information on the execution
command, number of repetitions and random seeds, runtime measurement
procedure, and computational hardware is provided in
\cref{app:technical_details}.

\subsection{Predictive performance}

\begin{figure*}[t]
  \centering
  \includegraphics[width=\textwidth]{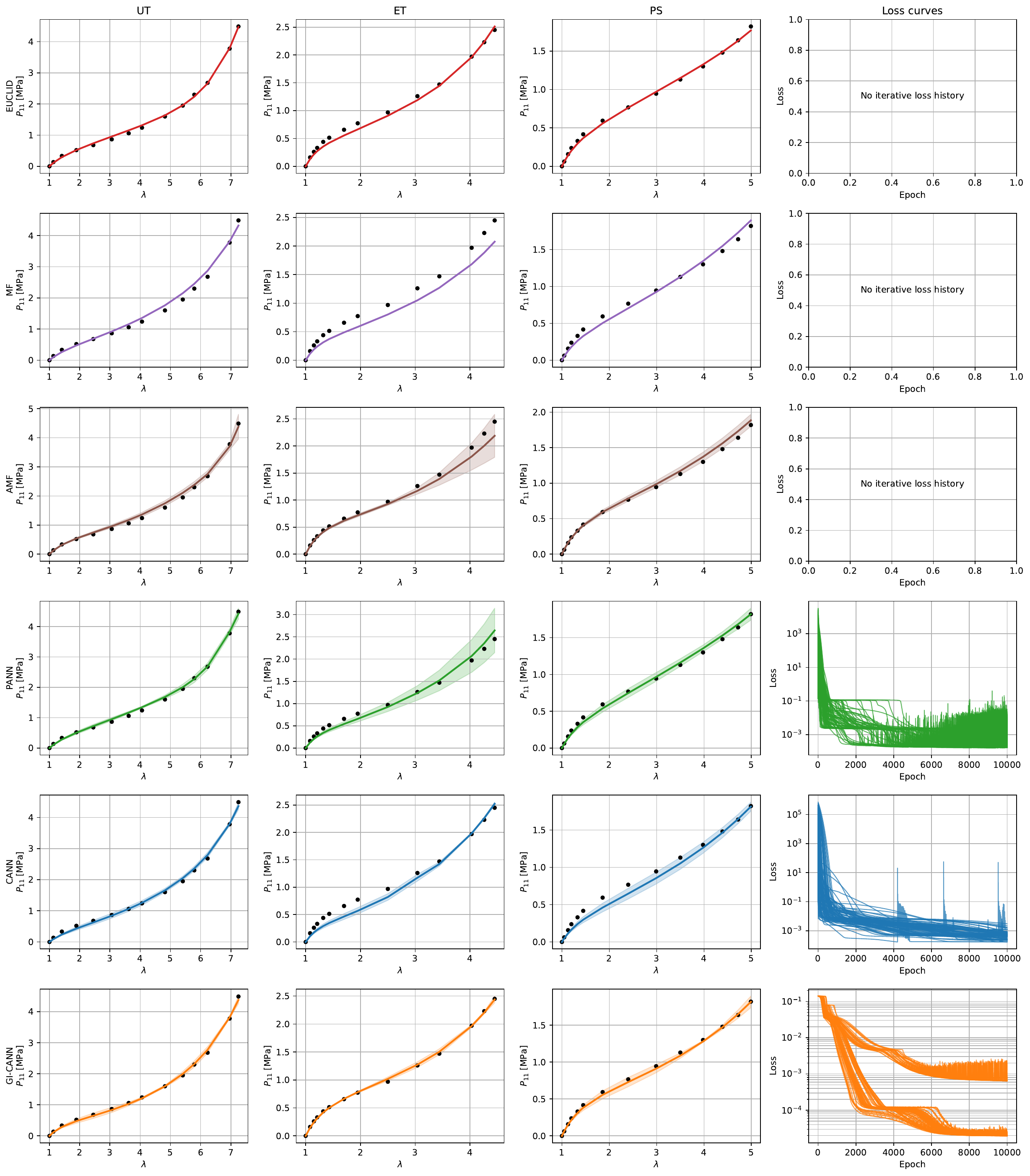}
  \caption{Predictive and training comparison of the six methods. Rows correspond to
  EUCLID, Material Fingerprinting (MF), Adaptive Material Fingerprinting (AMF),
  PANN, CANN, and GI-CANN. The first three columns show the first
  Piola--Kirchhoff stress component $P_{11}$ as a function of the stretch ratio
  $\lambda$ for uniaxial tension (UT), equibiaxial tension (ET), and pure shear
  (PS), respectively. Black markers denote the experimental reference data,
  solid lines denote the mean predictions, and shaded regions indicate one
  standard deviation. For AMF, the variation reflects the investigated 20
  hyperparameter configurations, whereas for PANN, CANN, and GI-CANN it reflects
  100 model initializations using different random seeds. The fourth column shows
  the individual training loss histories on a logarithmic scale. No iterative
  loss history is available for EUCLID, MF, or AMF.}
  \label{fig:prediction_method_grid}
\end{figure*}

All six methods reproduce the experimental stress-stretch responses with high accuracy across uniaxial tension, equibiaxial tension, and pure shear, as depicted in \cref{fig:prediction_method_grid,fig:best_models_comparison}. The predictions closely follow the experimental data over the complete stretch range, although the methods differ in their performance across deformation modes. The quantitative results in \cref{tab:metrics_summary} and \cref{fig:method_mse_summary,fig:method_r2_summary} show that all methods capture the dominant constitutive response.

EUCLID provides the highest accuracy for uniaxial tension and pure shear, with $R^2=0.9985$ and $0.9978$, respectively. GI-CANN performs particularly well for equibiaxial tension and reaches $R^2=0.9978$, whereas EUCLID reaches $R^2=0.9930$ for the same deformation mode. The differences between the methods become more apparent for equibiaxial tension, where MF, AMF, and PANN show larger deviations from the experimental response, see \cref{fig:prediction_method_grid}. In particular, PANN exhibits substantial variation across the different initializations.

Overall, the results show that the methods differ less in their ability to reproduce the overall constitutive response than in their robustness and accuracy for individual deformation modes. EUCLID performs particularly well despite its sparse representation, while GI-CANN provides consistently high accuracy across all three loading modes. The results also demonstrate that good agreement in the training modes does not automatically lead to identical performance in the testing pure shear mode.

\subsection{Best-performing models and model complexity}

We select one representative model for each method according to the minimum validation MSE in the testing pure shear mode. \cref{tab:best_models} summarizes the selected configurations, while \cref{fig:best_models_comparison} compares their predictions with the experimental data.
AMF provides the highest validation accuracy among the selected models, with $R^2=0.9996$. GI-CANN provides the highest training accuracy with $R^2=0.9996$ and maintains a validation value of $R^2=0.9993$. These results demonstrate that both methods can closely reproduce the experimental response, despite their fundamentally different model representations.

The selected models also differ substantially in complexity. MF produces the sparsest model with only two nonzero parameters. EUCLID requires four nonzero parameters and still provides a validation $R^2$ of $0.9978$. GI-CANN uses eight nonzero parameters, whereas CANN, AMF, and PANN use 12, 49, and 371 nonzero parameters, respectively, see \cref{tab:best_models}. PANN therefore requires substantially more parameters than the other methods without providing a corresponding increase in validation accuracy.

These results reveal a clear accuracy-complexity trade-off. For $N_a=20$, AMF provides the highest selected-model validation accuracy but requires a substantially larger representation than EUCLID or GI-CANN. We note, however, that the sparsity of AMF can be optimized through a simple hyperparameter search \citep{flaschel_adaptive_2026}. GI-CANN provides a particularly favorable compromise since it combines very high predictive accuracy with only eight nonzero parameters. EUCLID provides a different favorable compromise through its four-parameter representation and very low identification cost.

\subsection{Variability and optimization behavior}

\begin{figure*}[t]
  \centering
  \includegraphics[width=0.8\textwidth]{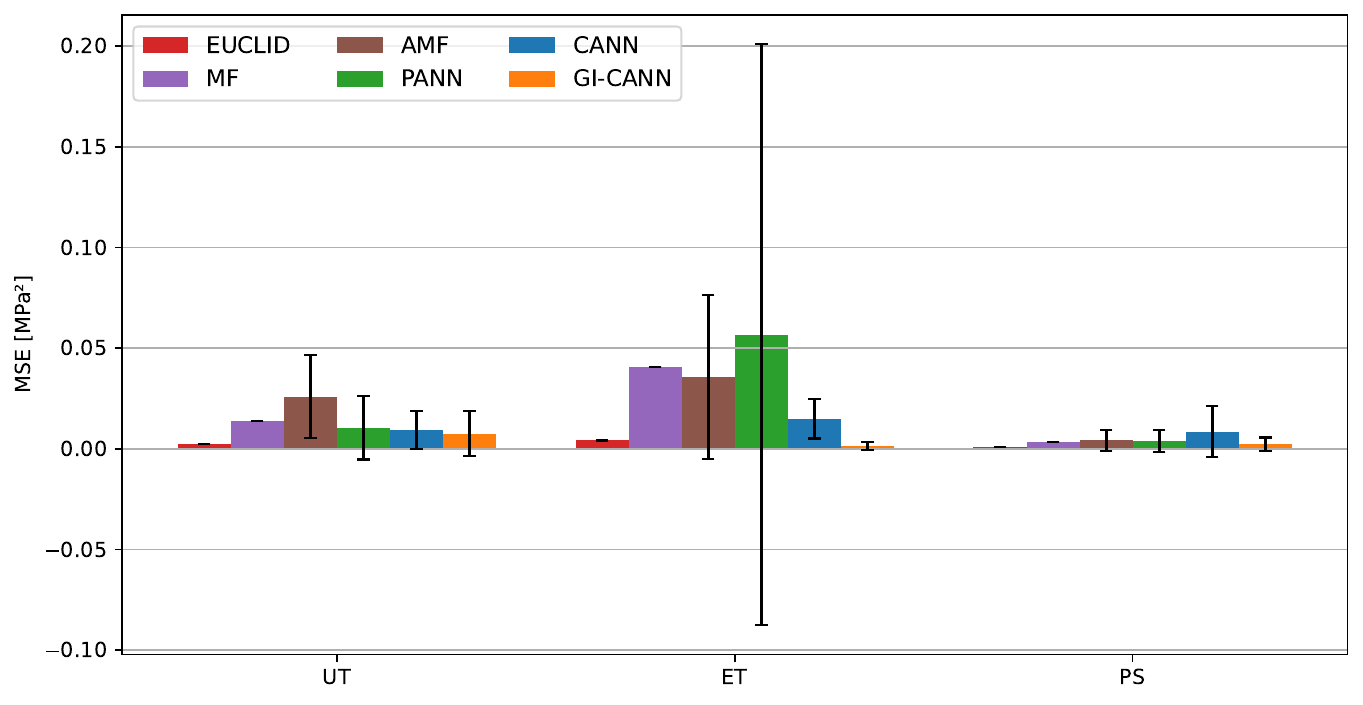}
  \caption{Mean squared error (MSE) of the six methods for uniaxial tension
  (UT), equibiaxial tension (ET), and pure shear (PS). Bars represent the
  arithmetic mean and error bars indicate one standard deviation. EUCLID and
  Material Fingerprinting (MF) each contribute one deterministic result;
  Adaptive Material Fingerprinting (AMF) comprises 20 combinations of the
  adaptive parameters $N_a$ and $s$; and PANN, CANN, and GI-CANN each comprise
  100 model initializations using different random seeds. Lower values indicate
  better predictive performance.}
  \label{fig:method_mse_summary}
\end{figure*}

\begin{figure*}[t]
  \centering
  \includegraphics[width=0.8\textwidth]{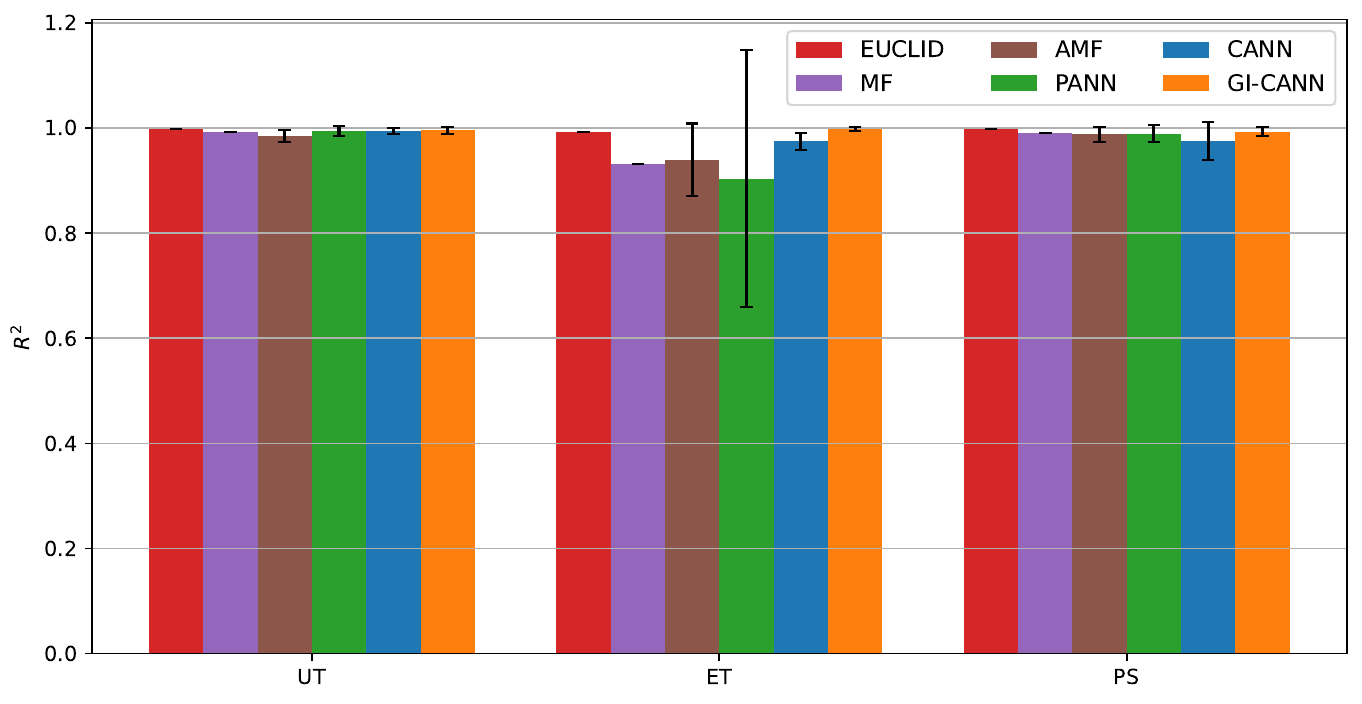}
  \caption{Coefficient of determination $R^2$ of the six methods for uniaxial
  tension (UT), equibiaxial tension (ET), and pure shear (PS). Bars represent
  the arithmetic mean and error bars indicate one standard deviation. EUCLID
  and Material Fingerprinting (MF) each contribute one deterministic result;
  Adaptive Material Fingerprinting (AMF) comprises 20 combinations of the
  adaptive parameters $N_a$ and $s$; and PANN, CANN, and GI-CANN each comprise
  100 model initializations using different random seeds. Values closer to one
  indicate better predictive performance.}
  \label{fig:method_r2_summary}
\end{figure*}

The repeated evaluations reveal clear differences in robustness between the methods. EUCLID and MF produce deterministic results for the selected settings, whereas AMF varies across the 20 investigated combinations of $N_a$~and~$s$. PANN, CANN, and GI-CANN vary across the 100 random initializations, see \cref{tab:metrics_summary,fig:method_mse_summary,fig:method_r2_summary}. For AMF, we note that we included $N_a=1$ for completeness, even though this reduces AMF to the classical MF with only one feature.

PANN shows the largest variability, particularly for equibiaxial tension. Its mean $R^2$ reaches only 0.9039 for this deformation mode, with a standard deviation of 0.2450. In contrast, GI-CANN reaches $R^2=0.9978\pm0.0035$ for the same deformation mode. CANN shows intermediate variability. The results, therefore, indicate substantially greater sensitivity to initialization for PANN than for GI-CANN.

\cref{fig:prediction_method_grid} illustrates this behavior through the prediction bands and training histories. PANN and CANN show pronounced differences between individual initializations, particularly for equibiaxial tension. GI-CANN also exhibits initialization-dependent convergence, but its prediction band remains comparatively narrow. This behavior suggests that the generalized invariant representation provides a more robust optimization landscape than the more flexible PANN architecture under the present benchmark conditions.

EUCLID, MF, and AMF do not use iterative gradient-based training during the reported identification stage. Their variability, therefore, does not originate from random network initialization. Instead, AMF exhibits variation across its adaptive hyperparameters, while MF and EUCLID depend on their predefined model or feature spaces.

\subsection{Computational cost of model identification}

\begin{figure*}[t]
  \centering
  \includegraphics[width=0.8\textwidth]{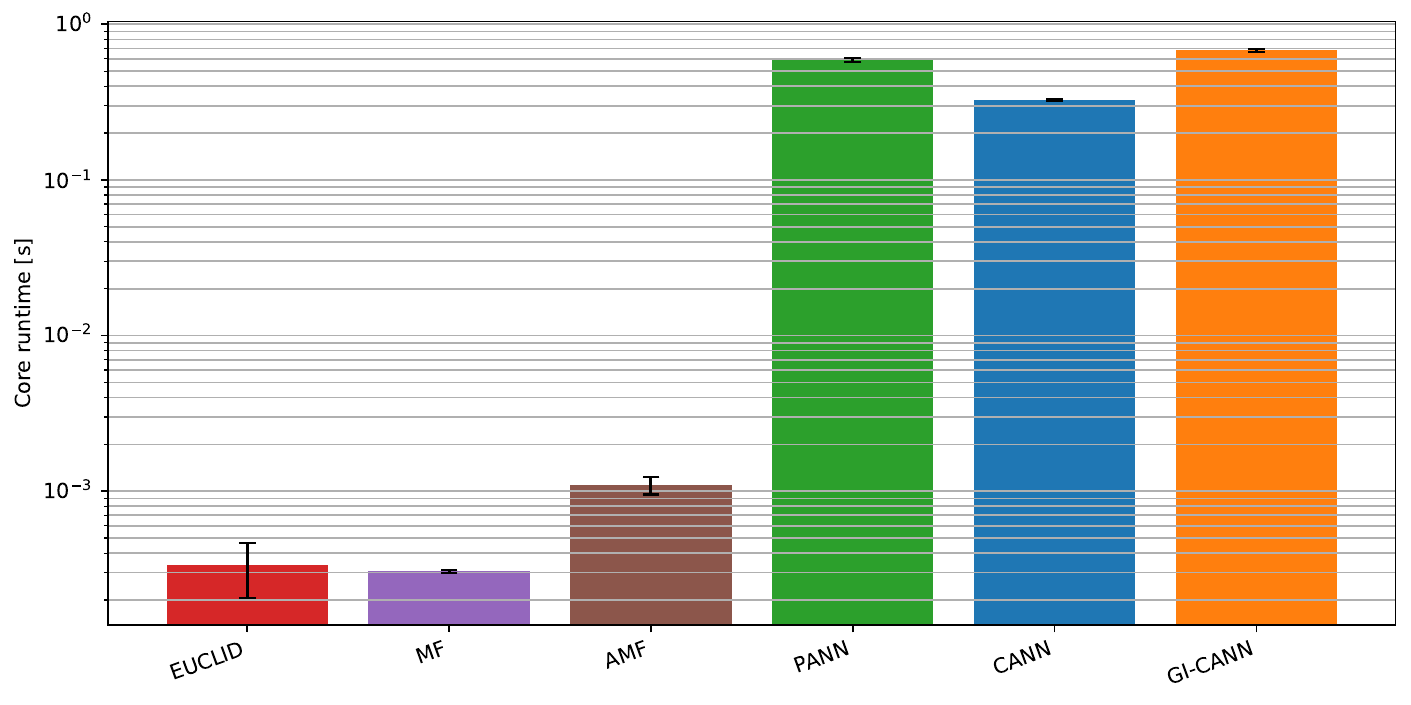}
  \caption{Method-specific core runtime of the six methods over 100 independent
  runs. Bars represent the arithmetic mean and error bars indicate one standard
  deviation. The core runtime comprises the Coordinate Descent solve for EUCLID, pattern
  recognition for Material Fingerprinting (MF), the adaptive loop for Adaptive
  Material Fingerprinting (AMF), and optimizer training after JAX warm-up for
  PANN, CANN, and GI-CANN. The runtime is reported in seconds on a logarithmic
  scale.}
  \label{fig:timing_summary}
\end{figure*}

The six methods show large differences in computational cost as shown in \cref{tab:timing_comparison,fig:timing_summary}. EUCLID and MF require less than $0.4$~ms for their core identification procedures, with mean core times of $0.337$~ms and $0.305$~ms, respectively. AMF requires $1.092$~ms and therefore remains substantially faster than the neural-network methods.

CANN, PANN, and GI-CANN require 324.539 ms, 588.495 ms, and 681.277 ms, respectively, for core optimization of 10,000 epochs. GI-CANN therefore requires more than three orders of magnitude more core computation than MF. The total identification time ranges from $0.491$~ms for EUCLID and $18.201$~ms for MF to approximately $0.4$--$0.8$~s for the three neural-network methods, see \cref{tab:timing_comparison}.

The results therefore separate two fundamentally different computational strategies. EUCLID, MF, and AMF perform sparse regression or database-based pattern recognition, whereas PANN, CANN, and GI-CANN solve nonlinear optimization problems. The latter approach provides greater flexibility in the constitutive representation but incurs a substantially higher identification cost.

The MF timing requires particular consideration because the database generation occurs offline. The reported core time represents only the online pattern-recognition step. Once the database exists, the same database can support repeated material-characterization tasks without repeating the database-generation process. This feature makes MF particularly attractive for applications that require repeated characterization. The computational advantage can become even more pronounced in the unsupervised setting, where MF can process heterogeneous full-field measurements without solving a continuous optimization problem \citep{flaschel_unsupervised_2026}. The present benchmark therefore focuses on the supervised online cost and does not include the one-time database-generation effort.

AMF follows the same offline database principle but adds an adaptive search over individual constitutive features. This additional search increases the online cost relative to MF, but AMF remains substantially faster than the neural-network methods while providing a more flexible constitutive representation.

\subsection{Computational cost of constitutive model evaluation}
\label{sec:computational_cost_model_evaluation}

\begin{figure*}[t]
  \centering
  \includegraphics[width=0.8\textwidth]{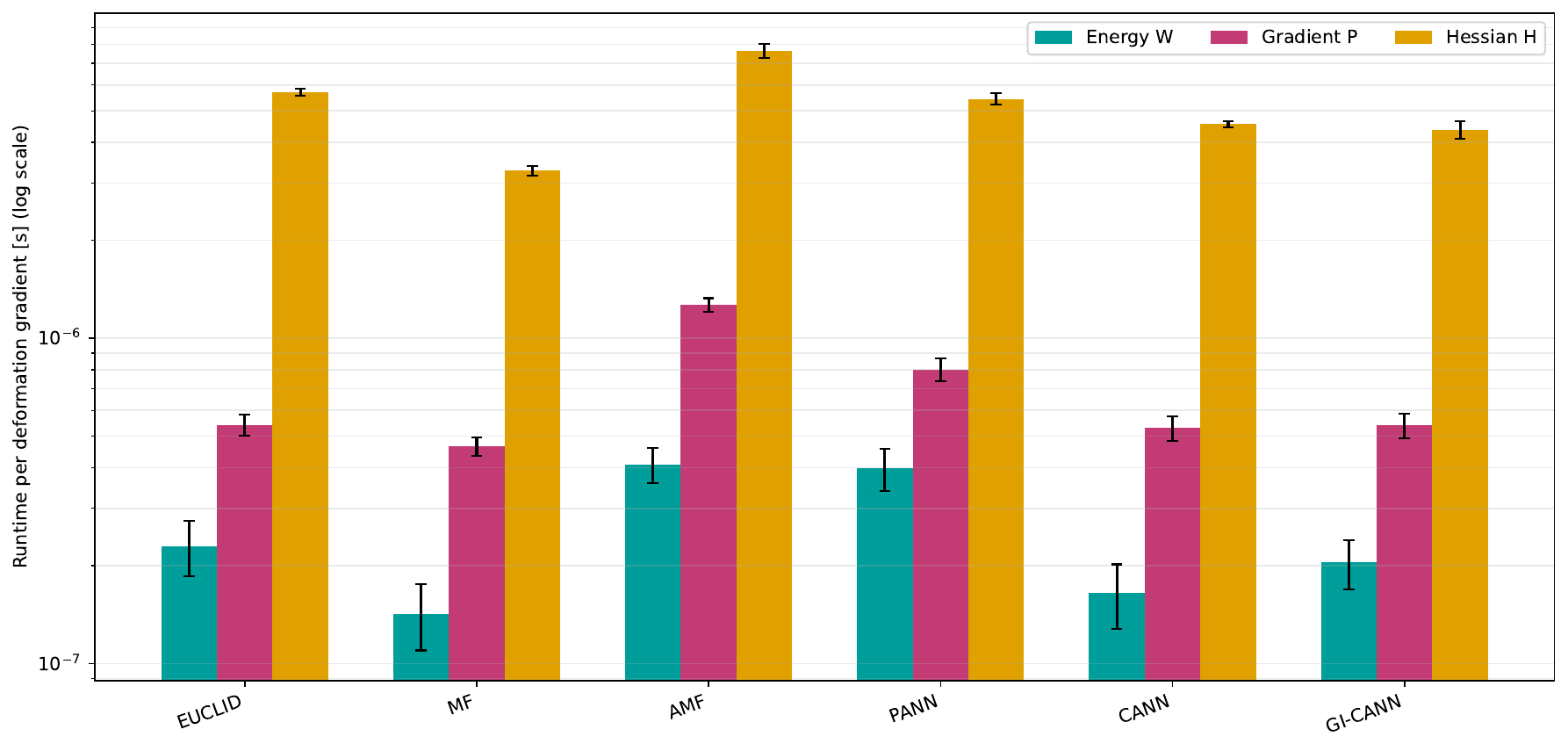}
  \caption{Runtime comparison for evaluating the strain energy density function
  $W$, its gradient $\bfP$, and its Hessian $\mathbb{H}$ for the six methods.
  Bars show the mean runtime per deformation gradient and error bars indicate
  one standard deviation over 100 repetitions after two warm-up runs.
  All evaluations were performed in double precision on the CPU using batches
  of 100 deformation gradients. The vertical axis uses a logarithmic scale.
  Compilation and other initialization steps are excluded from the reported
  runtimes. We note that for EUCLID and the neural network-based methods, we measured the evaluation times for the full model and did not exclude features that were identified as inactive during the inverse problem. Excluding such features could further reduce evauation times.}
  \label{fig:runtime_comparison}
\end{figure*}

The cost of model identification is independent of the cost of constitutive evaluation during a finite element simulation. Such simulations repeatedly evaluate the strain energy density, stress, and consistent tangent at every integration point. We therefore benchmark the evaluation of $W$, $\bfP$, and $\mathbb{H}$ for the six selected models, as shown in \cref{tab:benchmark_wph,fig:runtime_comparison}.

For this purpose, we implemented all six methods within a consistent computational framework and employed JAX automatic differentiation to compute both stresses and tangent operators. We compare the evaluation times of the best-performing model for each method.

MF provides the lowest evaluation time for all three quantities. It requires $0.151\,\mu\mathrm{s}$ per deformation gradient for $W$, $0.473\,\mu\mathrm{s}$ for $\bfP$, and $3.263\,\mu\mathrm{s}$ for $\mathbb{H}$. The Hessian requires the largest computation for every method, with AMF reaching $7.750\,\mu\mathrm{s}$ compared with $3.263\,\mu\mathrm{s}$ for MF, see \cref{tab:benchmark_wph}.

The results show that the number of nonzero parameters alone does not determine the cost of constitutive evaluation. PANN uses 371 nonzero parameters, but its evaluation times remain comparable to those of the other neural-network models. MF combines the smallest model with the lowest evaluation cost. This result highlights the importance of considering both model structure and implementation when assessing constitutive-model efficiency.

For EUCLID and the neural-network methods, we evaluate the complete selected model. We do not remove features or network components that become inactive during model identification. A reduced implementation could therefore further decrease the evaluation times, see \cref{fig:runtime_comparison}.

\subsection{Strain energy landscapes}

\begin{figure*}[t]
  \centering
  \includegraphics[width=\textwidth]{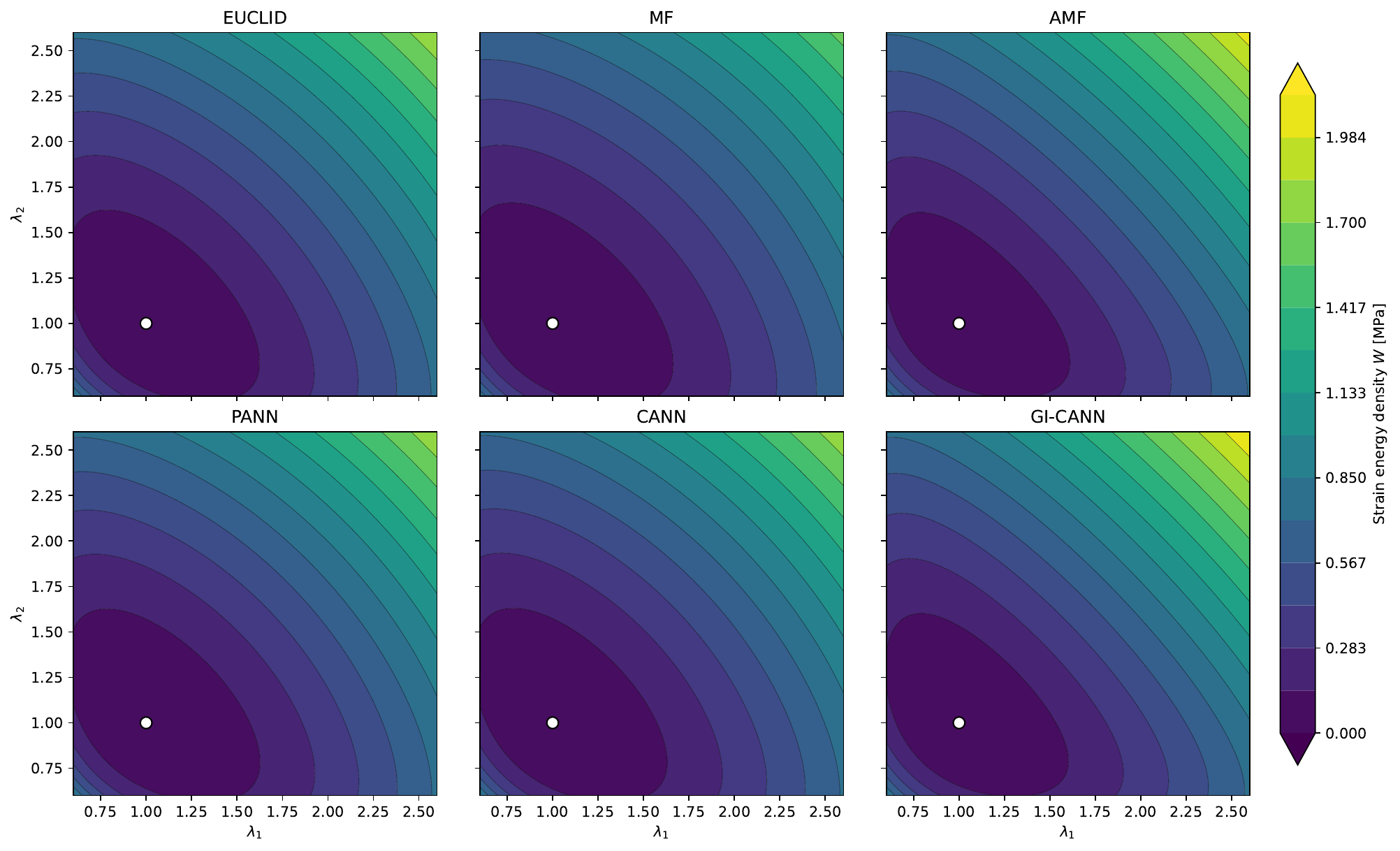}
  \caption{Strain energy density
  $W(\bfF)$ predicted by the selected
  best-performing models: EUCLID, Material Fingerprinting (MF), Adaptive
  Material Fingerprinting (AMF), PANN, CANN, and GI-CANN. Each panel shows
  $W$ as a function of the two independent principal stretches
  $\lambda_1$ and $\lambda_2$, with the third stretch determined by the
  incompressibility constraint
  $\lambda_3=(\lambda_1\lambda_2)^{-1}$. Identical contour levels and a common
  color scale are used in all six panels, enabling a direct comparison of the
  predicted energy landscapes. The marker at
  $(\lambda_1,\lambda_2)=(1,1)$ denotes the undeformed reference configuration.}
  \label{fig:strain_energy_isolines_grid}
\end{figure*}

We finally compare the constitutive functions beyond the deformation paths used for calibration. \cref{fig:strain_energy_isolines_grid} shows the strain energy density in principal-stretch space for the six selected models using identical contour levels and a common color scale. All models produce a smooth energy landscape with a minimum at the undeformed configuration $(\lambda_1,\lambda_2)=(1,1)$ and increasing energy with increasing deformation.

Despite their different mathematical representations, the six models produce qualitatively similar energy landscapes. EUCLID, AMF, PANN, CANN, and GI-CANN show closely related contour shapes over the displayed deformation range. MF produces a more distinct landscape, with systematically shifted contours relative to the other models. This difference agrees with the larger deviations of MF in the equibiaxial tension response shown in \cref{fig:best_models_comparison}.

The direct contour overlay in \cref{fig:strain_energy_isolines_overlay} provides a more detailed comparison of the predicted energy levels. The contours remain closely aligned at low and moderate energy levels, whereas the differences increase toward larger stretches and higher energy levels. These regions extend beyond the central part of the experimental data and therefore require greater extrapolation from the identified constitutive models. Interestingly, \cref{fig:strain_energy_isolines_overlay} reveals that models based on principal invariants (EUCLID, CANN, PANN) yield similar strain energy densities, while models based on principal stretches (AMF, GI-CANN) likewise exhibit similar strain energy densities. This underscores the importance of selecting appropriate model input features.

The individual landscapes in \cref{fig:strain_energy_isolines_grid} further show that sparse representations such as EUCLID and GI-CANN can reproduce energy surfaces that closely resemble those of the more complex adaptive and neural-network representations.

Despite their different mathematical representations, the six models produce qualitatively similar energy landscapes, see \cref{fig:strain_energy_isolines_overlay,fig:strain_energy_isolines_grid}. EUCLID, PANN, and CANN exhibit nearly identical contour shapes across the displayed stretch range, while AMF and GI-CANN show similarly close agreement. MF produces shifted contours relative to the other models, which are consistent with its larger deviations in the equibiaxial tension response.

\begin{table*}[t]
  \centering
  \caption{Computational performance of the six methods over
  100 independent runs. Timing results are reported in milliseconds as
  mean $\pm$ standard deviation. The method-specific core time comprises
  the Coordinate Descent solve for EUCLID, pattern recognition for Material
  Fingerprinting, the adaptive loop for Adaptive Material Fingerprinting,
  and optimizer training after JAX warm-up for PANN, CANN, and GI-CANN.
  Setup time comprises the remaining preprocessing and initialization
  steps outside the core computation. Bold values indicate the best
  result within the corresponding column.}
  \label{tab:timing_comparison}
  \small
  \setlength{\tabcolsep}{4.5pt}
  \begin{adjustbox}{max width=\textwidth}
  \begin{tabular}{@{}lrrr@{}}
    \toprule
    Method
      & \makecell{Core time\\(ms)}
      & \makecell{Setup time\\(ms)}
      & \makecell{Total time\\(ms)} \\
    \midrule
    EUCLID
      & $0.337 \pm 0.130$
      & $\mathbf{0.137 \pm 0.003}$
      & $\mathbf{0.491 \pm 0.133}$ \\
    MF
      & $\mathbf{0.305 \pm 0.006}$
      & $17.896 \pm 3.494$
      & $18.201 \pm 3.493$ \\
    AMF
      & $1.092 \pm 0.140$
      & $21.426 \pm 2.854$
      & $22.518 \pm 2.840$ \\
    PANN
      & $588.495 \pm 19.379$
      & $240.347 \pm 20.282$
      & $828.842 \pm 28.950$ \\
    CANN
      & $324.539 \pm 4.291$
      & $108.781 \pm 4.921$
      & $433.319 \pm 6.527$ \\
    GI-CANN
      & $681.277 \pm 14.110$
      & $117.587 \pm 12.135$
      & $798.863 \pm 18.259$ \\
    \bottomrule
  \end{tabular}%
  \end{adjustbox}
\end{table*}

\begin{table*}[t]
  \centering
    \caption{Predictive performance of the six methods for uniaxial tension (UT),
equibiaxial tension (ET), and pure shear (PS). Results are reported as arithmetic
mean $\pm$ standard deviation. The sample size $n$ has a method-specific meaning:
Adaptive Material Fingerprinting (AMF) comprises
20 combinations of the adaptive parameters $N_a$ and $s$, and PANN, CANN, and
GI-CANN each comprise 100 model initializations using different random seeds.
MSE, RMSE, and MAE denote the mean squared, root mean squared, and mean absolute
errors, respectively; NRMSE denotes the RMSE normalized by the range of the
reference data.
Bold values indicate the best mean performance within each loading mode.}
  \label{tab:metrics_summary}
  \small
  \setlength{\tabcolsep}{5pt}

  \begin{adjustbox}{max width=\textwidth}
  \begin{tabular}{@{}lcrrrrr@{}}
    \toprule
    Method & $n$ & MSE & RMSE & NRMSE & MAE & $R^2$ \\
    \midrule

    \multicolumn{7}{@{}l}{\textit{Uniaxial tension (UT)}} \\
    EUCLID & 1
      & \textbf{0.0025 $\pm$ 0}
      & \textbf{0.0501 $\pm$ 0}
      & \textbf{0.0112 $\pm$ 0}
      & \textbf{0.0389 $\pm$ 0}
      & \textbf{0.9985 $\pm$ 0}
      \\
    MF & 1
      & 0.0137 $\pm$ 0
      & 0.1172 $\pm$ 0
      & 0.0261 $\pm$ 0
      & 0.0962 $\pm$ 0
      & 0.9920 $\pm$ 0
      \\
    AMF & 20
      & 0.0258 $\pm$ 0.0207
      & 0.1443 $\pm$ 0.0722
      & 0.0321 $\pm$ 0.0161
      & 0.1033 $\pm$ 0.0588
      & 0.9850 $\pm$ 0.0120
      \\
    PANN & 100
      & 0.0103 $\pm$ 0.0156
      & 0.0874 $\pm$ 0.0520
      & 0.0195 $\pm$ 0.0116
      & 0.0698 $\pm$ 0.0392
      & 0.9940 $\pm$ 0.0091
      \\
    CANN & 100
      & 0.0093 $\pm$ 0.0095
      & 0.0887 $\pm$ 0.0385
      & 0.0198 $\pm$ 0.0086
      & 0.0725 $\pm$ 0.0295
      & 0.9946 $\pm$ 0.0055
      \\
    GI-CANN & 100
      & 0.0074 $\pm$ 0.0112
      & 0.0619 $\pm$ 0.0601
      & 0.0138 $\pm$ 0.0134
      & 0.0507 $\pm$ 0.0502
      & 0.9957 $\pm$ 0.0065
      \\

    \addlinespace
    \multicolumn{7}{@{}l}{\textit{Equibiaxial tension (ET)}} \\
    EUCLID & 1
      & 0.0041 $\pm$ 0
      & 0.0640 $\pm$ 0
      & 0.0261 $\pm$ 0
      & 0.0538 $\pm$ 0
      & 0.9930 $\pm$ 0
      \\
    MF & 1
      & 0.0407 $\pm$ 0
      & 0.2016 $\pm$ 0
      & 0.0823 $\pm$ 0
      & 0.1725 $\pm$ 0
      & 0.9310 $\pm$ 0
      \\
    AMF & 20
      & 0.0355 $\pm$ 0.0407
      & 0.1493 $\pm$ 0.1180
      & 0.0610 $\pm$ 0.0482
      & 0.0910 $\pm$ 0.0613
      & 0.9397 $\pm$ 0.0690
      \\
    PANN & 100
      & 0.0566 $\pm$ 0.1444
      & 0.1682 $\pm$ 0.1692
      & 0.0687 $\pm$ 0.0690
      & 0.1219 $\pm$ 0.1008
      & 0.9039 $\pm$ 0.2450
      \\
    CANN & 100
      & 0.0150 $\pm$ 0.0099
      & 0.1176 $\pm$ 0.0342
      & 0.0480 $\pm$ 0.0140
      & 0.0974 $\pm$ 0.0290
      & 0.9746 $\pm$ 0.0168
      \\
    GI-CANN & 100
      & \textbf{0.0013 $\pm$ 0.0020}
      & \textbf{0.0278 $\pm$ 0.0235}
      & \textbf{0.0114 $\pm$ 0.0096}
      & \textbf{0.0221 $\pm$ 0.0189}
      & \textbf{0.9978 $\pm$ 0.0035}
      \\

    \addlinespace
    \multicolumn{7}{@{}l}{\textit{Pure shear (PS)}} \\
    EUCLID & 1
      & \textbf{0.0008 $\pm$ 0}
      & \textbf{0.0277 $\pm$ 0}
      & \textbf{0.0152 $\pm$ 0}
      & \textbf{0.0221 $\pm$ 0}
      & \textbf{0.9978 $\pm$ 0}
      \\
    MF & 1
      & 0.0035 $\pm$ 0
      & 0.0596 $\pm$ 0
      & 0.0327 $\pm$ 0
      & 0.0503 $\pm$ 0
      & 0.9899 $\pm$ 0
      \\
    AMF & 20
      & 0.0042 $\pm$ 0.0052
      & 0.0504 $\pm$ 0.0420
      & 0.0277 $\pm$ 0.0231
      & 0.0378 $\pm$ 0.0308
      & 0.9880 $\pm$ 0.0149
      \\
    PANN & 100
      & 0.0037 $\pm$ 0.0055
      & 0.0516 $\pm$ 0.0319
      & 0.0283 $\pm$ 0.0175
      & 0.0423 $\pm$ 0.0262
      & 0.9895 $\pm$ 0.0158
      \\
    CANN & 100
      & 0.0085 $\pm$ 0.0126
      & 0.0805 $\pm$ 0.0450
      & 0.0442 $\pm$ 0.0247
      & 0.0671 $\pm$ 0.0409
      & 0.9758 $\pm$ 0.0359
      \\
    GI-CANN & 100
      & 0.0023 $\pm$ 0.0033
      & 0.0374 $\pm$ 0.0308
      & 0.0205 $\pm$ 0.0169
      & 0.0293 $\pm$ 0.0272
      & 0.9933 $\pm$ 0.0093
      \\

    \bottomrule
  \end{tabular}
  \end{adjustbox}
\end{table*}

\begin{table*}[t]
  \centering
  \caption{Best-performing configuration identified for each method. Models were
  selected exclusively by the minimum validation mean squared error (MSE) in the
  pure-shear (PS) mode within the comparison run set; test data were not used for
  model selection. The table reports the principal hyperparameters, the random seed
  for neural network models, training and validation MSE, training and validation
  coefficient of determination $R^2$, the method-specific core computation time,
  and the number of nonzero parameters.
  Bold values denote the best predictive result or lowest computational cost in the
  corresponding column.}
  \label{tab:best_models}
  \small
  \setlength{\tabcolsep}{4.5pt}
  \begin{adjustbox}{max width=\textwidth}
  \begin{tabular}{@{}lccrrrrrr@{}}
    \toprule
    Method
      & \makecell{Adaptive\\parameters}
      & Seed
      & \makecell{Training\\MSE}
      & \makecell{Validation\\MSE}
      & \makecell{Training\\$R^2$}
      & \makecell{Validation\\$R^2$}
      & \makecell{Core time\\(ms)}
      & \makecell{\# Parameters\\$\|\boldsymbol{\theta}\|_0$}
      \\
    \midrule
    EUCLID
      & --
      & --
      & $3.304\times10^{-3}$
      & $7.663\times10^{-4}$
      & 0.9973
      & 0.9978
      & 0.372
      & 4
      \\
    MF
      & --
      & --
      & $2.719\times10^{-2}$
      & $3.547\times10^{-3}$
      & 0.9781
      & 0.9899
      & \textbf{0.302}
      & \textbf{2}
      \\
    AMF
      & $N_a=20$, $s=0.25$
      & --
      & $1.119\times10^{-3}$
      & $\mathbf{1.524\times10^{-4}}$
      & 0.9991
      & \textbf{0.9996}
      & 6.000
      & 49
      \\
    PANN
      & --
      & 6
      & $3.329\times10^{-3}$
      & $8.186\times10^{-4}$
      & 0.9973
      & 0.9977
      & 581.311
      & 371
      \\
    CANN
      & --
      & 93
      & $3.535\times10^{-3}$
      & $8.925\times10^{-4}$
      & 0.9972
      & 0.9975
      & 320.947
      & 12
      \\
    GI-CANN
      & --
      & 39
      & $\mathbf{4.436\times10^{-4}}$
      & $2.412\times10^{-4}$
      & \textbf{0.9996}
      & 0.9993
      & 664.988
      & 8
      \\
    \bottomrule
  \end{tabular}%
  \end{adjustbox}
\end{table*}

\begin{table*}[t]
  \centering
  \caption{Computational benchmark for evaluating the strain energy density function
  $W$, its gradient $\bfP$, and its Hessian $\mathbb{H}$ for the six methods.
  Times are reported in microseconds per sample as mean $\pm$ standard deviation
  over 100 repetitions after two warm-up runs. Each repetition evaluated
  100 samples in a single chunk using double precision on the CPU.
  The reported times include one device synchronization per chunk and exclude
  compilation, parameter loading, sample generation, validation, and host result
  transfer. Bold values indicate the lowest mean evaluation time in the
  corresponding column.}
  \label{tab:benchmark_wph}
  \small
  \setlength{\tabcolsep}{4.5pt}
  \begin{adjustbox}{max width=\textwidth}
  \begin{tabular}{@{}lrrr@{}}
    \toprule
    Method
      & \makecell{Energy $W$\\($\mu\mathrm{s}$/sample)}
      & \makecell{Gradient $\bfP$\\($\mu\mathrm{s}$/sample)}
      & \makecell{Hessian $\mathbb{H}$\\($\mu\mathrm{s}$/sample)}
      \\
    \midrule
    EUCLID
      & $0.233 \pm 0.044$
      & $0.556 \pm 0.039$
      & $5.734 \pm 0.136$
      \\
    MF
      & $\mathbf{0.151 \pm 0.033}$
      & $\mathbf{0.473 \pm 0.030}$
      & $\mathbf{3.263 \pm 0.108}$
      \\
    AMF
      & $0.430 \pm 0.051$
      & $1.262 \pm 0.060$
      & $7.750 \pm 0.391$
      \\
    PANN
      & $0.428 \pm 0.059$
      & $0.827 \pm 0.064$
      & $5.496 \pm 0.218$
      \\
    CANN
      & $0.175 \pm 0.037$
      & $0.548 \pm 0.046$
      & $4.578 \pm 0.095$
      \\
    GI-CANN
      & $0.214 \pm 0.036$
      & $0.556 \pm 0.047$
      & $4.332 \pm 0.272$
      \\
    \bottomrule
  \end{tabular}%
  \end{adjustbox}
\end{table*}

\section{Conclusion}

We benchmark six data-driven constitutive modeling approaches on the classical Treloar rubber dataset under a common training and validation setup. AMF and GI-CANN provide the highest predictive accuracy among the selected models, while EUCLID and MF provide substantially lower identification costs and much sparser representations. GI-CANN combines high accuracy with a compact eight-parameter representation, whereas MF provides the lowest computational cost and the sparsest model. The benchmark therefore does not identify a single universally optimal method. Instead, the preferred approach depends on whether predictive accuracy, model sparsity, robustness, or computational efficiency receives the highest priority.

The results further show that constitutive model evaluation cost can differ substantially from model identification cost and therefore represents an important consideration for computational deployment. Overall, the benchmark highlights complementary strengths across the considered approaches and provides a quantitative basis for selecting data-driven constitutive models according to the requirements of a specific application.

\section*{Acknowledgments}

The authors acknowledge support from the European Research Council (ERC) Grant 101141626 DISCOVER funded by the European Union. 
Views and opinions expressed are, however, those of the authors only and do not necessarily reflect those of the European Union or the European Research Council Executive Agency. 
Neither the European Union nor the granting authority can be held responsible for them.

\section*{Data and code availability}

All datasets, numerical results, and implementation files required to reproduce the analyses presented in this work have been deposited in a public Zenodo repository. The archive also contains the implementations of the six investigated approaches together with the corresponding training and evaluation scripts. The repository is available at \href{https://doi.org/10.5281/zenodo.21915635}{https://doi.org/10.5281/zenodo.21915635}.

\section*{Disclosure of AI-assisted tools}

Generative artificial intelligence tools were used during the preparation of this manuscript to support language editing, improve clarity and readability, and assist with code development and debugging. All scientific content, methodological choices, numerical results, and conclusions were independently developed, verified, and approved by the authors, who take full responsibility for the content of the manuscript.

\clearpage
\bibliographystyle{elsarticle-harv}
\bibliography{bib_Moritz,bib_Hagen,bib_Denisa}

\appendix

\crefalias{section}{appendix}

\section{Technical details} \label{app:technical_details}

For all results, we use the following command

\begin{lstlisting}[language=bash, frame=none]
	python compare_all.py \
	--n-timing 100 --n-seeds 100 --euclid-lambdas 1e-5 \
	--study4-num-samples 100 --study4-repetitions 100
\end{lstlisting}

The option \texttt{--n-timing 100} performs 100 independent runtime measurements for each of the six constitutive model identification methods. The option \texttt{--n-seeds 100} evaluates the effect of neural-network initialization by training PANN, CANN, and GI-CANN with 100 different random seeds. The option \texttt{--euclid-lambdas 1e-5} restricts the EUCLID hyperparameter study to the regularization parameter $\alpha=10^{-5}$.

The option \texttt{--study4-num-samples 100} generates one common set of 100 admissible incompressible deformation gradients and evaluates all six models for the strain energy density $W$, the first Piola stress $\bfP=\frac{\partial W}{\partial\bfF}$, and the Hessian $\mathbb{H}=\frac{\partial^2 W}{\partial\bfF\partial\bfF}$. The option \texttt{--study4-repetitions 100} repeats the evaluation of this complete set 100 times for each model--quantity combination. We divide the total runtime of each repetition by 100 to obtain the runtime per deformation gradient. We then calculate the mean, median, and sample standard deviation across the 100 repetitions. We exclude compilation and warm-up executions from these statistics and record the cold-start time separately. For the hyperparameters and method-specific configuration values, we refer to \cref{tab:fixed_hyperparameters}.

We perform all computations on an Apple MacBook Pro with an Apple M4 processor that has a 10-core CPU with four performance cores and six efficiency cores and 24\,GB of unified memory. The system runs macOS~26.6. We perform all reported runtime measurements on the CPU backend.

Further, for the evaluation of the computational cost in \cref{sec:computational_cost_model_evaluation}, we compute the principal stretches as the singular values of the deformation gradient $\bfF$. Instead of explicitly performing a singular-value decomposition, we exploit that the squared singular values of $\bfF$ are the eigenvalues of the right Cauchy--Green tensor
\begin{equation}
    \bfC = \bfF^\mathsf{T}\bfF.
\end{equation}
Denoting the eigenvalues of $\bfC$ by $\mu_i$, the principal stretches follow as
\begin{equation}
    \lambda_i = \sqrt{\mu_i},
    \qquad i=1,2,3.
\end{equation}
Since $\bfC$ is symmetric, its eigenvalues are evaluated analytically using the closed-form trigonometric solution for a symmetric $3\times3$ matrix \citep{Kopp2008}. We first define
\begin{equation}
    q = \frac{1}{3}\operatorname{tr}(\bfC),
    \qquad
    \widehat{\bfC} = \bfC - q\bfI,
\end{equation}
and
\begin{equation}
    p
    =
    \sqrt{
        \frac{1}{6}
        \widehat{\bfC}:\widehat{\bfC}
    }.
\end{equation}
For $p>0$, we introduce
\begin{equation}
    \bfA = \frac{\widehat{\bfC}}{p},
    \qquad
    r = \frac{1}{2}\det(\bfA),
    \qquad
    \phi = \frac{1}{3}\arccos(r),
\end{equation}
where $r$ is numerically restricted to the interval $[-1,1]$. The largest and smallest eigenvalues of $\bfC$ are then obtained as
\begin{align}
    \mu_{\max}
    &=
    q + 2p\cos(\phi),
    \\
    \mu_{\min}
    &=
    q + 2p\cos\left(
        \phi+\frac{2\pi}{3}
    \right),
\end{align}
while the remaining eigenvalue follows from the trace constraint as
\begin{equation}
    \mu_{\mathrm{mid}}
    =
    3q-\mu_{\max}-\mu_{\min}.
\end{equation}
The principal stretches are subsequently computed as
\begin{equation}
    \lambda_1=\sqrt{\mu_{\max}},
    \qquad
    \lambda_2=\sqrt{\mu_{\mathrm{mid}}},
    \qquad
    \lambda_3=\sqrt{\mu_{\min}},
\end{equation}
such that $\lambda_1\geq\lambda_2\geq\lambda_3$. For nearly isotropic states, for which $p$ falls below a machine-precision-dependent threshold, we instead use
\begin{equation}
    \lambda_1=\lambda_2=\lambda_3=\sqrt{q},
\end{equation}
which avoids the numerical degeneracy of the trigonometric representation and ensures a well-defined evaluation of the undeformed reference configuration $\bfF=\bfI$.

\begin{table*}[t]
  \centering
  \caption{Fixed hyperparameters and method-specific configuration values. 
  The configurations were selected by
  minimizing the validation mean squared error in the pure-shear (PS) mode without
  using test data. For PANN, CANN, and GI-CANN, the learning rate, AdamW optimizer
  settings, and number of epochs were fixed across
  the studies. Except for the method-specific learning rates,
  the default Optax settings of AdamW were used. The random seeds are therefore not
  listed as hyperparameters. For Adaptive Material
  Fingerprinting (AMF), the values shown correspond to the selected configuration,
  whereas the aggregate AMF performance statistics comprise 20 investigated
  combinations of the adaptive parameters $N_a$ and $s$.}
  \label{tab:fixed_hyperparameters}
  \small
  \setlength{\tabcolsep}{7pt}
  \begin{adjustbox}{max width=\textwidth}
  \begin{tabular}{@{}ll@{}}
    \toprule
    Hyperparameter & Value \\
    \midrule
    \multicolumn{2}{@{}l}{\textit{EUCLID}} \\
    Optimization method & Coordinate Descent \\
    $\alpha$ & $1\times10^{-5}$ \\
    $\theta_{\mathrm{th}}$ & 0 \\
    \addlinespace
    \multicolumn{2}{@{}l}{\textit{Material Fingerprinting (MF)}} \\
    Material database & Hyperelastic incompressible (HEI) \\
    \addlinespace
    \multicolumn{2}{@{}l}{\textit{Adaptive Material Fingerprinting (AMF)}} \\
    Material database & Hyperelastic incompressible isotropic adaptive (HEIIA) \\
    $N_a$ & [1, 3, 5, 10, 20] \\
    $s$ & [0.25, 0.5, 0.75, 1.0] \\
    \addlinespace
    \multicolumn{2}{@{}l}{\textit{Physics-Augmented Neural Network (PANN)}} \\
    Optimizer & AdamW (Optax) \\
    Weight constraints & Projected Gradient Descent \\
    Learning rate & $1\times10^{-2}$ \\
    Remaining optimizer settings & Optax defaults \\
    Number of epochs & 10,000 \\
    Hidden layer widths & $[16,16]$ \\
    Output dimension & 1 \\
    Hidden layer activation & Softplus \\
    Output activation & Softplus \\
    \addlinespace
    \multicolumn{2}{@{}l}{\textit{Constitutive Artificial Neural Network (CANN)}} \\
    Optimizer & AdamW (Optax) \\
    Weight constraints & Projected Gradient Descent \\
    Learning rate & $1\times10^{-4}$ \\
    Remaining optimizer settings & Optax defaults \\
    Number of epochs & 10,000 \\
    \addlinespace
    \multicolumn{2}{@{}l}{\textit{Generalized Invariant-based CANN (GI-CANN)}} \\
    Optimizer & AdamW (Optax) \\
    Weight constraints & Projected Gradient Descent \\
    Learning rate & $1\times10^{-3}$ \\
    Remaining optimizer settings & Optax defaults \\
    Number of epochs & 10,000 \\
    \bottomrule
  \end{tabular}
  \end{adjustbox}
\end{table*}

\end{document}